\documentclass[draftcls,onecolumn,journal]{IEEEtran}

\usepackage{caption}
\usepackage{cite}
\usepackage{amsmath,amssymb,amsfonts}
\usepackage{graphicx}
\usepackage{textcomp}
\usepackage{xcolor}
\usepackage{hyperref}
\usepackage{amsbsy}
\usepackage{amsthm, mathtools}
\usepackage{epsfig}
\usepackage{subcaption}
\usepackage{epstopdf}
\usepackage{blindtext}
\usepackage{comment}
\usepackage{diagbox}
\usepackage{tikz}
\usepackage{lipsum}

\usepackage{algorithm,algpseudocode}
\algnewcommand{\Inputs}[1]{
	\State \textbf{Inputs:}
	\Statex \hspace*{\algorithmicindent}\parbox[t]{.8\linewidth}{\raggedright #1}
}
\algnewcommand{\Initialize}[1]{%
	\State \textbf{Initialize:}
	\Statex \hspace*{\algorithmicindent}\parbox[t]{.8\linewidth}{\raggedright #1}
}
\allowdisplaybreaks[1]

\providecommand{\abs}[1]{\lvert#1\rvert}
\renewcommand{\b}[1]{\ensuremath{\mathbf{#1}}} 
\newcommand{\bs}[1]{\ensuremath{\boldsymbol{#1}}} 
\newcommand{\Ex}[1]{\ensuremath{\mathbb{E}[#1]}}  
\newcommand{\norm}[1]{\ensuremath{\left\|#1\right\|}} 
\newcommand{\tb}[1]{\ensuremath{\tilde{\mathbf{#1}}}} 
\newcommand{\leqtext}[1]{\ensuremath{\stackrel{\text{#1}}{\leq}}} 
\newcommand{\geqtext}[1]{\ensuremath{\stackrel{\text{#1}}{\geq}}} 

\newcommand{\lfrom}[1]{{\leqtext{\eqref{#1}}}}
\newcommand{\gfrom}[1]{{\geqtext{\eqref{#1}}}}

\def \i {{\b{1}}}
\def \a {{\b{a}}}

\def \g {{\b{g}}}
\def \s {{\b{s}}}
\def \u {{\b{u}}}
\def \v {{\b{v}}}
\def \w {{\b{w}}}
\def \x {{\b{x}}}
\def \y {{\b{y}}}
\def \z {{\b{z}}}

\def \xc {{\check{\x}}}
\def \Qc {{\check{\b{Q}}}}

\def \T {{\mathsf{T}}}		

\def \ux {{\underline{\b{x}}}}
\def \ug {{\underline{\b{g}}}}
\def \us {{\underline{\b{s}}}}
\def \uy {{\underline{\b{y}}}}

\def \tQ {{\tb{Q}}}

\def \Ps {{$\mathcal{P}_s$}}

\def \A {{\b{A}}}
\def \B {{\b{B}}}
\def \C {{\b{C}}}
\def \D {{\b{D}}}
\def \E {{\b{E}}}
\def \F {{\b{F}}}
\def \G {{\b{G}}}
\def \H {{\b{H}}}
\def \I {{\b{I}}}
\def \L {{\b{L}}}
\def \M {{\b{M}}}
\def \O {{\b{0}}}
\def \P {{\b{P}}}
\def \Q {{\b{Q}}}
\def \R {{\b{R}}}
\def \S {{\b{S}}}
\def \V {{\b{V}}}
\def \W {{\b{W}}}
\def \Z {{\b{Z}}}

\def \xks {{\x^{k,\star}}}

\def \cE {{\mathcal{E}}}

\def \cG {{\mathcal{G}}}
\def \cH {{\mathcal{H}}}

\def \cN {{\mathcal{N}}}
\def \cO {{\mathcal{O}}}

\def \cV {{\mathcal{V}}}

\def \nab {{\bs{\nabla}}}

\def \Reg {{\mathbf{Reg}}}

\def \Rn {{\mathbb{R}}}
\def \Nn {{\mathbb{N}}}

\theoremstyle{definition}

\newtheorem{theorem}{Theorem}
\newtheorem{lemma}{Lemma}
\newtheorem{corollary}{Corollary}

\newtheorem{assumption}{}

\begin{document}
	
	\title{Optimized Gradient Tracking for \\
		Decentralized Online Learning \author{Shivangi Dubey Sharma and Ketan Rajawat, \textit{Member, IEEE}}}

	\maketitle
	
	\begin{abstract}
		This work considers the problem of decentralized online learning, where the goal is to track the optimum of the sum of time-varying functions, distributed across several nodes in a network. The local availability of the functions and their gradients necessitates coordination and consensus among the nodes. We put forth the Generalized Gradient Tracking (GGT) framework that unifies a number of existing approaches, including the state-of-the-art ones. The performance of the proposed GGT algorithm is theoretically analyzed using a novel semidefinite programming-based analysis that yields the desired regret bounds under very general conditions and without requiring the gradient boundedness assumption. The results are applicable to the special cases of GGT, which include various state-of-the-art algorithms as well as new dynamic versions of various classical  decentralized algorithms. To further minimize the regret, we consider a condensed version of GGT with only four free parameters. A procedure for offline tuning of these parameters using only the problem parameters is also detailed. The resulting  optimized GGT (oGGT) algorithm not only achieves improved dynamic regret bounds, but also outperforms all state-of-the-art algorithms on both synthetic and real-world datasets. 
	\end{abstract}
	
	
	\section{Introduction}
	Decentralized systems form the bedrock of the Big Data era. The immense surge in the data volumes renders centralized storage and processing impractical. Privacy and anonymity concerns when dealing with sensitive data, such as that sourced from hospitals, further discourage centralization. Recent years have witnessed the  rise of large-scale decentralized learning systems that achieve scalability and circumvent privacy issues by allowing the data to be stored locally \cite{nedic2009distributed,ram2010distributed,nedic2011asynchronous,jakovetic2014fast}. 
	
	In this work, we focus on decentralized learning in dynamic environments, where new data points are continuously streaming in at every node. As the distribution of the streaming data might change over time, online learning algorithms must be deployed to deal with the problem of \emph{conceptual drift}. For instance, in applications such as environmental monitoring and autonomous navigation, the data distribution may depend on the current environment, e.g., on the amount of ambient light. To prevent the learned models from becoming obsolete, it becomes necessary to update them regularly \cite{yan2012distributed, hosseini2013online}. 
	
	Mathematically, the decentralized online learning problem can be written as that of minimizing the sum of time-varying functions distributed among several nodes. The component functions, each corresponding to a data point stored at a node, are differentiable and the nodes are only allowed to exchange algorithm parameters or gradients among themselves. In this context, decentralized online learning algorithms have been well-studied in the last decade \cite{yan2012distributed,nedic2009distributed ,ram2010distributed,nedic2011asynchronous,hosseini2013online,jakovetic2014fast,lee2017stochastic, yi2020distributed, oakamoto2023distributed}. The performance of these algorithms is usually calculated in terms of the  (static) regret, which measures the cumulative loss incurred by the learner over a horizon, as compared to that incurred by a static model constructed from the complete data. More recently, researchers have shifted focus to \emph{dynamic regret}, where the cumulative loss of the learner is compared against that incurred by a dynamic model that is obtained by minimizing the loss at every time instant. While dynamic regret bounds are generally more pessimistic than the static regret bounds, the dynamic regret is a more appropriate measure in time-varying scenarios where a static benchmark lacks a meaningful interpretation. 
	
	Classical decentralized optimization algorithms, such as Distributed Gradient Descent (DGD), are well-known to be suboptimal due to error accumulation at every iteration \cite{yuan2016convergence}. To mitigate these errors, various algorithms proposed in the literature utilize two classes of strategies: (a) combining consecutive DGD updates resulting in correction term(s), as in EXTRA \cite{shi2015extra} and NIDS \cite{li2019decentralized}, and (b) locally tracking the global gradient, as in DIGing and ATC-GT \cite{qu2017harnessing, nedic2017achieving}. These gradient tracking (GT) algorithms have shown superior performance in closely tracking the global gradient, thereby achieving linear convergence rates to the exact optimum, comparable to that of the centralized gradient descent. 
	
	Dynamic regret performance of decentralized online optimization algorithms has been well-studied \cite{shahrampour2018distributed, nazari2021adaptive}. Gradient tracking has also been implemented and analyzed in this context, and is known to yield improved regret bounds \cite{zhang2019distributed, zhang2020distributed, carnevale2022gtadam}. Of these, the works in \cite{zhang2020distributed, carnevale2022gtadam} achieve state-of-the-art regret rates without assuming gradient boundedness or compactness of the domain. However, not all decentralized optimization algorithms have been adapted or analyzed for the dynamic setting at hand. Likewise, no generalized or unified algorithms for decentralized online optimization exist, in contrast to the static generalized optimization algorithms such as \cite{alghunaim2022unified, alghunaim2020decentralized}.

	Recognizing the existing gaps in decentralized online learning, we introduce the Generalized Gradient Tracking (GGT) algorithm that unifies a number of existing decentralized online optimization algorithms. At the same time, the GGT algorithm seeks to achieve the state-of-the-art dynamic regret matching the effectiveness of centralized optimization algorithms. For instance, algorithms proposed in \cite{zhang2019distributed, zhang2020distributed} are special cases of GGT and have regret bounds that are worse than those of GGT. Further, we construct dynamic versions of the decentralized algorithms proposed in \cite{shi2015extra, li2019decentralized, hong2017prox, qu2017harnessing, xu2015augmented, scutari2019distributed}, which are also special cases of GGT. The performance of GGT is analyzed using a robust semidefinite programming-based approach, first proposed for classical static optimization in \cite{sundararajan2017robust}. We modify the approach to allow time-varying smooth local functions $f_i^k$ such that $f_i^k(\x) - \frac{1}{2}\x^\T\M_i\x$ is convex for positive semidefinite $\M_i$ but still require the overall function $\sum_{i=1}^n f_i^k$ to be strongly convex. The proposed approach allows us to numerically ascertain the exact dynamic regret bound that would result from using a given set of algorithm parameters. Subsequently, we construct a condensed version of GGT with only four free parameters. By adjusting these free parameters to minimize the regret bound, we obtain the optimized GGT (oGGT), which not only attains the smallest possible dynamic regret but also outperforms all existing state-of-the-art decentralized online learning algorithms. 
	
	The remaining sections are organized as follows: Sec. \ref{secprob} provides the problem formulation and introduces the dynamic regret measure. Additionally, we review some of the state-of-the-art decentralized optimization and learning algorithms that are related to the current work. The proposed GGT algorithm is developed in Sec.  \ref{secunif} and a number of concrete examples are provided. The regret bounds for the GGT algorithm as well as the procedure to obtain the oGGT algorithm are provided in Sec. \ref{secregret}. Sec. \ref{secnumanalysis} details the numerical experiments on both synthetic and real data sets. Finally, Sec. \ref{secconclusion} concludes the paper. 
	
	\textit{Notations}: We denote vectors (matrices) using lowercase (uppercase) bold font letters. For a vector $\x$, we denote its transpose by $\x^\T$ and its $i$-th element by $[\x]_i$. Likewise, the $(i,j)$-th component of $\A$ is given by $A_{ij}$. Identity and all-zero matrices is denoted by $\I_n$  and $\O_{m \times n}$, respectively, but we will also use $\O_n$ in place of $\O_{n \times n}$ for the sake of brevity. Further, we will drop the subscripts of $\I$  and $\O$ if the size is clear from the context. The $n \times 1$ all-one vector is denoted by $\i$. The Hadamard and Kronecker products are denoted using binary operators $\odot$ and $\otimes$, respectively.  The block diagonal matrix constructed by keeping the matrices $\A_1$, $\A_2$, $\ldots$, along the main diagonal while making the other entries zero is denoted by $\text{Diag}(\{\A_i\})$. For a vector $\y$, $\norm{\y}$ dentoes its $\ell_2$ (Euclidean) norm, $\Vert \y \Vert_\P\coloneqq(\y^\T \P\y)^{1/2}$ denotes its $\P$-norm, and  $\norm{\y}_\infty$ denotes its  $\ell_\infty$-norm. The notation $\A\succ 0$ implies that matrix $\A$ is a positive definite matrix. Maximum and minimum eigenvalues of $\A$ are represented by $\lambda_{\max}(\A)$ and $\lambda_{\min}(\A)$, respectively, and cond($\A$) denotes the ratio of $\lambda_{\max}(\A)$ and $\lambda_{\min}(\A)$. The set of neighbors of node $i$ is denoted by $\cN_i$.

	\section{Problem Formulation}
	\label{secprob}
	Consider a network of $n$ agents or nodes communicating over a fixed undirected graph $\mathcal{G}=(\cV,\cE)$, where $\cV$ is the set of nodes and $\cE$ is a set of edges or links. An edge $ (i,j) \in \cE$ represents a communication link between nodes $i$ and $j$. Further, let $\E \in \Rn^{n \times n}$ denote the degree matrix of the graph and $\Q\in \Rn^{\abs{\cE} \times n}$ be its incidence matrix. For an edge $e=(i,j) \in \cE$, where $j > i$ without loss of generality, we have 
	\begin{align}
		Q_{e,\ell} = \begin{cases}
			1	& \ell = i, \\
			-1 & \ell = j, \\
			0 & \text{ otherwise. }
		\end{cases}
	\end{align}
	
	We consider a streaming scenario where data points arrive sequentially over time and each data point corresponds to a loss function. Examples of loss functions include regression and classification losses in machine learning, the likelihood function in parameter estimation, the risk function in finance, and time-series cross-validation loss in forecasting. Denoting the global convex function at time $k \in \Nn$ by $f^k:\Rn^d\rightarrow \Rn$, the network of nodes seek to cooperatively track the optimum $\xks\coloneqq\arg\min_{\x} f^k(\x)$ over time. We consider a canonical distributed setting, where an agent $i$ only receives a local convex function $f_i^k:\Rn^d\rightarrow \Rn$ at each $k \geq 1$ \cite{nedic2015decentralized,johansson2007simple,georgiadis2006resource}, and the global function is given by 
	\begin{align}
		f^k(\x) = \frac{1}{n}\sum_{i=1}^n f_i^k(\x). \label{fav}
	\end{align}
	The agents are not aware of the global function $f^k$ and must therefore communicate among each other to determine and track $\xks$. 
	
	We approach this problem from the lens of online learning: at time $k$, agent $i$ decides upon its action $\x_i^k$ and subsequently, an adversary reveals $\nabla f_i^k(\x_i^k)$. Such an adversarial model allows us to consider arbitrary and potentially non-stationary data streams, such as those encountered in finance, sensor networks, and social media. Given the time-varying nature of the objective function, the tracking performance of the proposed algorithm is quantified by the so-called dynamic regret of the multi agent system given by \cite{shahrampour2018distributed} 
	\begin{equation} \label{dyreg}
		\hspace{-2mm}\Reg^K \!\!\coloneqq \tfrac{1}{n}\sum_{j=1}^n \Reg_j^K \!\!=  \tfrac{1}{n}\sum_{j=1}^n \sum_{k=1}^K \left(f^k(\x_j^k)-f^k(\xks)\right),
	\end{equation}
	where $\Reg_j^K\coloneqq \sum_{k=1}^K \left(f^k(\x_j^k)- f^k(\xks)\right)$ is the dynamic regret due to agent $j$,  $f^k(\x_j^k)\coloneqq \frac{1}{n} \sum_{i=1}^n f_i^k(\x_j^k)$, and $f^k(\xks)\coloneqq \frac{1}{n} \sum_{i=1}^n f_i^k(\xks)$.  
	As evident from \eqref{dyreg}, the functions $\{f^k_i\}$ as well as the optimum $\xks$ are both time-varying. The setting is different from that of static regret, where the goal is to compare against a static benchmark $\x^\star$. Indeed, the dynamic regret is not sublinear in general, but can, however, be bounded in terms of the problem complexity, quantified by different path variation metrics. Here, we consider two such metrics: the cumulative path length \cite{mokhtari2016online,zhang2017improved,bedi2018tracking,zhang2018adaptive,lesage2019online} given by
	\begin{align}
		C_{K,p} &\coloneqq \sum_{k=2}^K \norm{\x^{k,\star}-\x^{k-1,\star}}^p,  \label{ck}
	\end{align}
	and similar to \cite{dixit2019online,chiang2012online, chiang2013beating,zhang2019distributed,zhang2020distributed}, the cumulative gradient difference given by 
	\begin{align}
		D_{K,p} & \coloneqq \sum_{k=2}^K\norm{\nab^{k,\star}-\nab^{k-1,\star}}^p, \label{dk}
	\end{align}
	where $\nab^{k,\star} = [(\nabla f_1^k(\x^{k,\star}))^\mathsf{T}, \dots , (\nabla f_n^k(\x^{k,\star}))^\mathsf{T}]^\mathsf{T} \in \Rn^{nd}$, and $p\geq 1$. The subsequent analysis yields bounds that depend either on $C_{K,1}$ and $D_{K,1}$ (represented henceforth by $C_K$ and $D_K$, respectively) or on $C_{K,2}$ and $D_{K,2}$.
	
	In order to minimize the dynamic regret, we put forth a class of distributed first order methods inspired from similar algorithms in the static context. Given problem parameters, the hyper-parameters of the proposed class of algorithms can be chosen to minimize the regret term, resulting in the oGGT algorithm, which also exhibits superior empirical performance. 
	
	\subsection{Review of decentralized optimization algorithms}
	We begin with reviewing the state-of-the-art decentralized optimization algorithms, where the goal is to solve the static problem:
	\begin{align}\label{static}\tag{\Ps}
		\x^\star = \arg\min \sum_{i=1}^n f_i(\x),
	\end{align}
	in a decentralized manner. These can be considered as solving the batch version of the problem at hand, where the entire data is available a priori and incorporated into $f_i$. As we shall see in Sec. \ref{secunif}, the proposed dynamic decentralized algorithms are based largely on these static algorithms.
	
	Various algorithms developed in this domain includes gradient methods \cite{yuan2016convergence}, distributed subgradient methods \cite{nedic2009distributed,ram2010distributed,matei2011performance}, primal-dual methods \cite{bianchi2015coordinate,duchi2011dual}, Alternating Direction Method of Multipliers (ADMM) \cite{boyd2011distributed,iutzeler2015explicit,ling2015dlm}, and accelerated gradient descent \cite{jakovetic2014fast, chen2012fast}.

	The classical gradient-based method to solve \eqref{static} in a distribution fashion is the Decentralized Gradient Descent (DGD) algorithm \cite{tsitsiklis1986distributed, nedic2009distributed}, which allows each node to maintain a local copy $\x_i^k$. The update at each agent 
	takes the form:
	\begin{equation}
		\label{DGDupdate}
		\x_i^{k+1}=\sum_{j=1}^{n} W_{i,j}\x_j^k-\eta^k \nabla f_{i}(\x_i^k),
	\end{equation} 
	where, $\eta^k$ is the step-size parameter and $\W$ is a symmetric doubly stochastic matrix with $W_{ij} > 0$ if and only if $(i,j)\in \cE'$ where $\cE' = \cE \cup \{(i,i)\}_{i=1}^n $. The DGD algorithm has been well-studied \cite{yuan2016convergence} and applied to a wide variety of settings \cite{matei2011performance, nedic2018distributed}.
	
	The Exact First Order Algorithm (EXTRA) improves upon DGD and its variants, yielding better rates and allowing step-size ($\eta$) values that do not depend on the connectivity properties of $\cG$ \cite{shi2015extra}. The EXTRA updates take the form
	\begin{equation} \label{extraupdate}
		\begin{split} 
			\x_{i}^{k+1} &=\x_{i}^{k}+\sum\limits_{j=1}^{n}W_{ij}\x_{j}^{k}-\sum\limits_{j=1}^{n}\V_{ij}\x_{j}^{k-1} -\eta\left(\nabla f_{i}(\x_{i}^{k})-\nabla f_{i}(\x_{i}^{k-1})\right), 
		\end{split}
	\end{equation}
	where $\W$ and $\V$ are symmetric doubly stochastic matrices. As in DGD, $W_{ij} = V_{ij} = 0$ if $i = j$ or if nodes $i$ and $j$ are not immediate neighbors. The Network Independent Step-Size (NIDS) algorithm offers further flexibility by allowing each node to choose its own network-independent step-sizes, by further mixing the gradient tracking terms of neighboring nodes in the update equation\cite{li2019decentralized}. 
	
	More recently, the gradient tracking has emerged as a generic approach to accelerating decentralized optimization algorithms, and has found application in a variety of settings, including those involving directed and time-varying graphs \cite{sun2017distributed, scutari2019distributed, lu2019gnsd}. The idea is to track the network-wide gradient using the iterates \cite{nedic2017achieving, qu2017harnessing}:
	\begin{align}\label{gt1}
		\g_i^{k+1}&=\sum_{j=1}^{n} W_{ij}\g_j^k+\bigg{(}\nabla f_{i}(\x_{i}^{k+1})-\nabla 	f_{i}(\x_{i}^k) \bigg{)},
	\end{align}
	and subsequently use the tracked gradient $\g_i^k$ to carry out DGD-like updates
	\begin{align}\label{gt2}
		\x_i^{k+1}&= \sum_{j=1}^{n} W_{ij}\x_j^k- \eta \g_i^k,
	\end{align}
	resulting in the DIGing algorithm \cite{nedic2017achieving}.
	
	Instead of adding a consensus term to the update equations, one could also introduce local variables $\{\x_i\}_{i=1}^n$ and impose 	 constraints $\x_i = \x_j$ for all $(i,j)\in \cE$ in \eqref{static}. Collecting the vectors $\{\x_i\}_{i=1}^n$ into a super-vector $\check{\x} \in \Rn^{nd}$, Problem \eqref{static} can equivalently be written as
	\begin{equation}
		\label{aim_cons_compact}
		\underset{\xc\in \Rn^{nd}}{\min} f(\xc) \quad \text{s. t.} \quad  \Qc\xc=0,
	\end{equation}
	where $\Qc \coloneqq \Q \otimes \I_d$, with $\Q$ being the adjacency matrix. The problem in \eqref{aim_cons_compact} can now be solved in a decentralized fashion using various augmented Lagrangian, dual, and primal-dual approaches \cite{boyd2011distributed,hong2017prox,chang2020distributed}.
	
	For streaming scenario with independent identically distributed (i.i.d.) data, stochastic variants of these algorithms been developed. Let $f_i^k(\cdot)$ denote the loss function corresponding to the random data point arriving at time $k$ so that $f_i(\x) = \Ex{f_i^k(\x)}$. The stochastic version of DGD, referred to as the Distributed Stochastic Gradient Descent (DSGD) \cite{ram2010distributed}, entails using the stochastic gradient $\nabla f_i^k(\x_i^k)$ at time $k$ instead of the regular gradient in \eqref{DGDupdate}. Likewise, the $D^2$ algorithm is the stochastic variant of the gradient tracking algorithm, and achieves a better dependence on the data variance \cite{tang2018d}. Finally, the Gradient tracking-based Non-convex Stochastic Decentralized (GNSD) algorithm outperforms $D^2$ by using a more general mixing matrix.

	\begin{table*}[ht] 
		\centering
		\caption{State-of-the-art decentralized online learning algorithms } \label{table:regret_rate}
		\begin{center}
			\begin{tabular}{|l|c|c|l|c|}
				\hline
				\multicolumn{1}{|c|}{\textbf{Algorithm}} & 
				\multicolumn{1}{l|}{\textbf{\begin{tabular}[c]{@{}l@{}}Constrained/\\ Unconstrained\end{tabular}}} & \multicolumn{1}{l|}{\textbf{Function}}                                            & \textbf{Regret Rate} & \textbf{Special case of GGT}\\ \hline
				DMD \cite{shahrampour2018distributed} & Constrained                                                                                                    &  \begin{tabular}[c]{@{}c@{}}Convex\\ Bounded Gradient \end{tabular} & $\cO \left(\sqrt{K (1 + C_K)  }\right)$  & No                    \\ \hline
				DDAG\cite{nazari2021adaptive} & Constrained                                                                                                    &  \begin{tabular}[c]{@{}c@{}}Convex\\ Bounded Gradient\end{tabular}        &$\cO \left(\sqrt{K(1+\log K)}+\sqrt{K} \left(1+ C_K\right)+1\right)$    & No                       \\ \hline
				DOG \cite{wu2022decentralized}  & Unconstrained                                                                                                       & \begin{tabular}[c]{@{}c@{}}Convex\\ Smooth\\ Bounded Gradient\end{tabular} & $\cO \left(\sqrt{K \left(1+ C_K\right) }\right)$   &No                    \\ \hline
				DOO-GT\cite{zhang2019distributed}  & Unconstrained                                                                                                 &  \begin{tabular}[c]{@{}c@{}}Strongly Convex\\ Smooth \\ Bounded Gradient\end{tabular}                    &$\cO \left( 1+ C_K +V_K \right)$   & Yes                       \\ \hline
				D-OCO \cite{zhang2020distributed}    & Unconstrained                                                                                               &  \begin{tabular}[c]{@{}c@{}}Strongly Convex\\ Smooth\end{tabular}                    &$\cO \left( 1+ C_K + V_K^\star \right)$   & Yes                    \\ \hline
				GTAdam \cite{carnevale2022gtadam}   & Unconstrained                                                                                             &  \begin{tabular}[c]{@{}c@{}}Strongly Convex\\ Smooth\end{tabular}                    & $\cO \left( 1+ C_K + V_K^m + \sqrt{C_K + V_K^m} \right)$       &No              \\ \hline
				GGT (This Work)                & Unconstrained                                                                                   & 
				\begin{tabular}[c]{@{}c@{}}Strongly Convex\\ Smooth\end{tabular}                    &$\cO \left( 1+ \min\{C_K +D_K, C_{K,2} +D_{K,2}\} \right)$            &$-$          \\ \hline
			\end{tabular}
		\end{center}
	\end{table*}

	\subsection{Review of decentralized online learning algorithms}
	Decentralized algorithms for non-stationary or online settings have also been explored. The decentralized mirror descent (DMD) was proposed in \cite{shahrampour2018distributed} to solve constrained time-varying optimization problems, 
	and achieves the dynamic regret rate of $\mathcal{O}(\sqrt{K (1 + C_K)})$. Distributed Dynamic Adaptive Gradient (DDAG) is a dynamic and adaptive variant of DGD that utilizes the momentum to achieve a regret rate of Distributed Dynamic Adaptive Gradient algorithm is $\cO \left(\sqrt{K(1+\log K)}+\sqrt{K} \left(1+ C_K \right)+1\right)$, which is better than that of DMD when the gradient vector is sparse.
	The regret rate can be further improved when the feasible region is compact, as shown in \cite{dixit2020online,nazari2022dadam,lesage2020dynamic,eshraghi2022improving}. However, algorithms designed for the compact setting cannot be applied to the unconstrained problem at hand. The Decentralized Online Gradient (DOG) algorithm, which is the online variant of the DSGD \cite{ram2010distributed}, achieves the dynamic regret of $\cO (\sqrt{K (C_K+1)} )$ for the unconstrained setting \cite{wu2022decentralized}. The Distributed Online Optimization with Gradient Tracking (DOO-GT) was the first work to use gradient tracking in decentralized online learning, and achieved an improved rate of  $\cO(1+C_K+V_K)$ \cite{zhang2019distributed} as compared to \cite{wu2022decentralized,shahrampour2018distributed,nazari2021adaptive} for strongly convex functions. The DOO-GT algorithm eliminates the dependence of regret on time step $K$ and instead depends on the cumulative gradient difference $V_K =\sum_{k=1}^{K} \norm{\nab^k- \nab^{k-1}}_\infty$ where $\nab^{k}=[(\nabla f_1^k(\x_1^{k}))^\T,...,(\nabla f_n^k(\x_n^{k}))^\T]^\T$.  
	
	The recently proposed distributed online convex optimization algorithm (D-OCO) is a slight modification of the DOO-GT algorithm, but does not require the compactness assumption, which is crucial in DOG and DOO-GT \cite{zhang2020distributed}. D-OCO achieves a regret rate of $\mathcal{O}(1+C_K+V_K^\star)$ without requiring the compactness assumption, with $V_K^\star=\sum_{k=1}^{K} \norm{\nab^k(\x^{k-1, \star})- \nab^{k-1}(\x^{k-1, \star})}$ where $\nab^{k}(\x)\coloneqq[(\nabla f_1^k(\x))^\T,\ldots,(\nabla f_n^k(\x))^\T] ^\T$. Observe here that $V_K^\star$ is different from the cumulative gradient difference $D_K$ in \eqref{dk}. In particular, $V_K^{\star}$, utilizes the difference in the consecutive function gradients, evaluated at a specific point $\x^{k-1,\star}$. In contrast, $D_K$ measures the change in the consecutive function gradient evaluations at their respective optima. While $D_K$ and $V_K^\star$ are not directly comparable, we note that there exist situations where $D_K = 0$ while $V_K^\star > 0$, such as in the case when all the nodal functions have the same optimum value. The Gradient Tracking with Adaptive momentum estimation (GTAdam) algorithm incorporates momentum into the DOO-GT to achieve 
	a regret rate of $\cO \left( 1+ C_K + V_K^m + \sqrt{C_K + V_K^m} \right)$, where $V_K^m = \sum_{k=0}^{K-1}\underset{i}{\max}\,\underset{\x \in \Rn^d}{\max} \norm{\nabla f_i^{k+1}(\x) -  \nabla f_i^{k}(\x)}$ \cite{carnevale2022gtadam}. More recently, a mirror-descent-based algorithm DOMD-MADGC with multiple consensus steps per-iteration has been shown to achieve $\cO(1+C_K)$ for strongly-convex, smooth functions with bounded gradient \cite{eshraghi2022improving}. The algorithm however requires a compact domain which is not necessarily possible for the unconstrained setting considered here. Further, the need for multiple rounds of communication at every iteration may not always be possible, and hence not considered here. 
	
	The present work provides a unified algorithm which subsumes several gradient tracking-based algorithms, including D-OCO and DOO-GT, and achieves the dynamic regret bound of $\cO(1+ \min\{C_K + D_K, C_{K,2} + D_{K,2}\})$ without requiring the feasible region to be compact. 
	
	\section{Unified Algorithm}
	\label{secunif}
	This section details the proposed GGT framework, which brings together the strengths of numerous existing algorithms and unifies them for decentralized online convex optimization. It is well-known from \cite{yuan2016convergence} that the errors incurred by the DGD algorithm prevent it from converging linearly to the optimum. To mitigate these errors, various decentralized optimization algorithms utilize two classes of strategies: (a) combining consecutive DGD updates resulting in correction term(s), as in EXTRA and NIDS, and (b) locally tracking the global gradient, as in DIGing and ATC-GT. The proposed GGT updates incorporate both these strategies using various parameter matrices $\cH \coloneqq \{\H_{ij}^{\ell}\}_{(i,j)\in\cE',\ell \in \{1,\ldots,8\}}$ where $\cE' = \cE \cup \{(i,i)\}_{i=1}^n$. The different versions of the algorithm, corresponding to the specific choices of these matrices, will be introduced subsequently. For ease of exposition, we write down the updates for $k=0$ and $k \geq 1$ separately.
	
	In each iteration of GGT, each node maintains two variables: namely, $\x_i^k$ (the local iterate) and $\s_i^k$ (the auxiliary variable). Each update step of GGT involves two distinct stages: a generalized consensus and descent step to update the local iterate and a generalized tracking step to monitor the auxiliary variable. The generalized consensus and descent step involves consensus (similar to DGD) as well as progress along the corrected descent direction (similar to GT) to update the local iterate $\x_i^k$ for each node $i$ at iterate $k \geq 1$ and takes the following form:
	\begin{align}\label{xi-up}
		\x_i^{k} &= \sum_{(i,j) \in \cE'} \H^{(1)}_{ij}\x_j^{k-1} + \sum_{(i,j) \in \cE'} \H^{(2)}_{ij} \s_j^k.
	\end{align}
	The update in \eqref{xi-up} involves the consensus term, which is similar to DGD, and the descent term depending on $\s_i^k$, which can be viewed as a locally tracked copy of the global gradient. Observe that if the second term in \eqref{xi-up} is removed, then it would amount to classical consensus (averaging) without any progress. Further, non-GT variants, namely EXTRA, NIDS, and PGPDA, do not have the consensus step at all, i.e., $\H^{(1)} = \mathbf{0}$.
	Having obtained $\x_i^{k}$ and taken the corresponding action, each node queries the oracle, which reveals the value of $\nabla f_i^{k}(\x_i^{k})$. Next, the generalized tracking step at node $i$ updates the auxiliary variable, providing the corrected descent direction that each node must follow locally to reach the global optimum. This corrected descent direction is updated dynamically and takes the form:
	\begin{align}\label{si-up}
		\s_i^{k+1} &= \sum_{(i,j)\in \cE'} \H^{(3)}_{ij}\x_j^{k-1} + \sum_{(i,j)\in \cE'} \H^{(4)}_{ij} \s_j^k +  \sum_{(i,j)\in \cE'} \H^{(5)}_{ij} \left(\nabla f_j^{k-1}(\x_j^{k-1})  - \nabla f_j^{k}(\x_j^{k})\right).
	\end{align}
	The update in \eqref{si-up} involves the standard gradient tracking terms as well as correction terms depending on the previous iterates at neighboring nodes $\x_j^{k-1}$. Without the $\x_j^{k-1}$ term in \eqref{si-up}, the update becomes a classical gradient tracking step, as in ATC-GT and DIGing. Addition of $\x_j^{k-1}$ from the neighboring nodes and previous iterations helps reduce the error in $\s_i^{k}$. The correction term is not present in any of the GT variants, i.e., $\H^{(3)} = \mathbf{0}$ for these algorithms.
	
	The algorithm is initialized with arbitrary $\x_{i}^0$, which is then utilized to calculate $\nabla f_i^{0}(\x_i^{0})$ and the initial value of $\s_{i}^0$ should be such that it satisfies
	\begin{align}
		0 = \sum_{(i,j)\in \cE'} \H^{(6)}_{ij}\x_j^{0} + \sum_{(i,j)\in \cE'} \H^{(7)}_{ij}\s_j^0  +  \sum_{(i,j)\in \cE'} \H^{(8)}_{ij} \nabla f_j^{0}(\x_j^{0}). \label{initlztn}
	\end{align}
	The updates of GGT can be seen as linear combinations of iterates $\x_j^{k-1}$, auxiliary variables $\s_j^{k-1}$, and gradient differences $\nabla f_j^{k-1}(\x_j^{k-1}) - \nabla f_j^k(\x_j^k)$, across neighboring nodes. The weights of these linear combinations are kept as general parameter matrices $\cH$. In general, these weights can be tuned depending on the problem parameters and the network. Further $\cH$ allows us to define various algorithms present in literature as a special cases of the Generalized Gradient Tracking (GGT).
	
	These updates can be compactly written by stacking the quantities $\x_i^k$, $\s_i^k$, and $\nabla f_i^k(\x_i^k)$ into $nd$-dimensional super vectors, $\ux^k$, $\us^k$, and $\nab^k$, respectively. Defining $\H^{(\ell)} \in \Rn^{nd \times nd}$ such that its $(i,j)$-th block matrix $[\H^{(\ell)}]_{ij} = \H_{ij}^{(\ell)}$ for $(i,j)\in\cE'$ and $\mathbf{0}_d$ otherwise, for each $\ell \in \{1, \ldots, 8\}$, the updates in \eqref{xi-up}-\eqref{initlztn} can be written as
	\begin{equation}
		\begin{aligned}\label{compact} 
			\ux^{k} &= \H^{(1)}\ux^{k-1} + \H^{(2)} \us^{k-1}, \\
			\us^{k} &= \H^{(3)}\ux^{k-1} + \H^{(4)} \us^{k-1} + \H^{(5)}(\nab^{k-1}-\nab^{k}),
		\end{aligned}
	\end{equation}
	\begin{align} \label{initialz_cmpct0}
		\H^{(6)}\ux^0 + \H^{(7)} \us^0 + \H^{(8)}\nab^0 &= \O_{d \times 1},
	\end{align}
	for $\H^{(6)}, \H^{(7)}, \H^{(8)} \in \Rn^{d \times nd}$. It can be verified that if we choose $\H^{(6)}, \H^{(7)}$, and $\H^{(8)}$ such that
	\begin{subequations}
		\label{Mequalities}
		\begin{align}
			\H^{(6)} \H^{(1)} + \H^{(7)} \H^{(3)} &= \H^{(6)}, \\
			\H^{(6)} \H^{(2)} + \H^{(7)} \H^{(4)} &= \H^{(7)}, \\
			\H^{(7)} \H^{(5)} &= \H^{(8)}, 
		\end{align}
	\end{subequations}
	and utilize the update equation \eqref{xi-up}-\eqref{initialz_cmpct0} recursively, then a condition similar to \eqref{initialz_cmpct0} can be written for all $k \geq 1$, i.e.,
	\begin{equation} \label{initialz_cmpct}		
		\H^{(6)}\ux^{k-1} + \H^{(7)} \us^{k-1} + \H^{(8)}\nab^{k-1} = \O_{d \times 1}.
	\end{equation}
	For the detailed proof of equation \eqref{initialz_cmpct}, refer to Appendix~\ref{pr:eq:initialz_cmpct}. 
	Henceforth, we will assume that the matrices in $\cH$ are chosen so as to satisfy \eqref{Mequalities}. The full algorithm is summarized in Algorithm \ref{Ualgo}.
	\begin{algorithm}
		\caption{Generalized Gradient Tracking (GGT) Algorithm}\label{Ualgo}
		\begin{algorithmic}[1]
			\State	\textbf{Require} $\cH$
			\State \textbf{Initialize} $\ux^0$
			\State \textbf{Calculate} $\us^0$ so as to satisfy \eqref{initialz_cmpct0}
			\For{$k = 1,2,3,...$ and each user $i$}
			\State \textbf{Update} $\x_i^{k}$ using \eqref{xi-up}
			\State \textbf{Perform} action  $\x^{k}_i$ and observe $\nabla f_i^{k}(\x_i^{k})$
			\State \textbf{Update} $\s^{k}_i$  using  \eqref{si-up}
			\EndFor
		\end{algorithmic}
	\end{algorithm}
	
	Let us consider a more concrete form of the proposed algorithm by relating it to the EXTRA algorithm \cite{shi2015extra}, initially proposed for the static setting. Using the notation $\s_i^{k-1} = \x_i^{k}$, the updates in \eqref{extraupdate} can be expressed as
	\begin{equation}\label{extraupdate2}
		\begin{aligned}
			\x_i^{k} &= \s_i^{k-1}, \\
			\s_i^{k} &= \s_i^{k-1} + \sum_{(i,j)\in\cE} W_{ij} \s_j^{k-1} - \sum_{(i,j)\in\cE} V_{ij} \x_j^{k-1}-\eta\left(\nabla f_i(\x_{i}^{k})-\nabla f_i(\x_{i}^{k-1})\right). 
		\end{aligned}
	\end{equation}
	where recall that $W_{ij}$ and $V_{ij}$ are entries of symmetric doubly stochastic mixing matrices $\W$ and $\V$ respectively, such that $\V \succ 0$, $\text{null}\{\W-\V\} = \text{span}\{\mathbf{1}\}$ and $\frac{\I+\W}{2}\succeq \V \succeq \W$. The similarity between the forms of the updates in \eqref{extraupdate2} and \eqref{xi-up}-\eqref{si-up} motivates the dynamic EXTRA (D-EXTRA) algorithm which takes the form:
	\begin{align} \label{D-EXTRAupdate}
		\x_{i}^{k+1} &=\x_{i}^{k}+\sum\limits_{j=1}^{n}W_{ij}\x_{j}^{k}-\sum\limits_{j=1}^{n}V_{ij}\x_{j}^{k-1}-\eta\left(\nabla f_i^k(\x_{i}^{k})-\nabla f_i^{k-1}(\x_{i}^{k-1})\right). 
	\end{align}
	Similar to EXTRA, the algorithm is initialized with arbitrary $\x_i^0$ and $\x_i^1 = \x_i^0 - \nabla f^0_i(\x_i^0)$ for all $1\leq i\leq n$. It can be seen that D-EXTRA is a special case of the proposed algorithm, and corresponds to parameter choices given in the first column of Table \ref{newformtable}. The choice of matrices for D-EXTRA also satisfies \eqref{Mequalities}. The implementation of the D-EXTRA algorithm is summarized in Algorithm \ref{D-EXTRA}. Where, $\cN_i$ represents the set of neighboring nodes of node $i$. As the proposed algorithm is a decentralized algorithm, so in the absence of central node, each node communicates only with its neighboring nodes.
	\begin{algorithm}
		\caption{D-EXTRA}\label{D-EXTRA}
		\begin{algorithmic}[1]
			\State	\textbf{Require} $\eta$, $\W$, $\V$
			\State \textbf{Initialize} $\x_i^0$ and $\x_i^1=\sum_{j=1}^{n}[\W]_{ij}\x_j^0-\eta \nabla f_i^0(\x_i^0) \quad \forall i$
			\For {$k = 1,2,...$ and each user $i$ } 
			\State \textbf{Perform} action $\x_i^k$
			\State \textbf{Observe} gradient $\nabla f_{i}^k(\x_{i}^k)$ at $\x_i^k$
			\State \textbf{Communicate} neighbors' local models $\x_j^k$, $j \in \cN_i$ 
			\State \textbf{Compute} update action as in (\ref{D-EXTRAupdate})
			\EndFor
		\end{algorithmic}
	\end{algorithm}
	
	Proceeding along similar lines, we propose the dynamic NIDS (D-NIDS) \cite{li2019decentralized},  dynamic PG-PDA (D-PGPDA) \cite{hong2017prox}, dynamic gradient estimation (D-GE) \cite{qu2017harnessing}, dynamic Adapt Then Combine-GT (D-ATC-GT) \cite{xu2015augmented}, dynamic gradient tracking (D-GT)  \cite{scutari2019distributed}, which is equivalent to dynamic OCO (D-OCO) \cite{zhang2020distributed}, and DOO-GT proposed in \cite{zhang2019distributed}, which is also a special case of GGT. The parameter matrices for these variants are summarized in Table \ref{newformtable} and satisfy \eqref{Mequalities}. 
	
	For the DPGDA algorithm, $\tilde{\E}=:\E \otimes \I_d$ and $\tQ = \Q \otimes \I_d$, so that $\tilde{\Z}=\frac{1}{2}\tilde{\E}^{-1}(\L^+-\L^-)$  where $\L^-=\tQ^\T \tQ\in \Rn^{nd \times nd}$ and $\L^+=2\tilde{\E}-\tQ^\T \tQ \in \Rn^{nd \times nd}$. The implementation-ready form of the updates are provided in the Appendix \ref{updates}; see \eqref{D-NIDSupdate},\eqref{D-PGPDA},\eqref{DQuLi}, \eqref{D-ATC-GT}, and \eqref{D-GT}. 
	
	Recently in \cite{alghunaim2022unified}, a general (static) decentralized optimization algorithm is proposed with the following update rule
	\begin{equation}\label{ga}
		\begin{aligned}
			\ux^{k+1}&=\tilde{\A} \tilde{\C}\ux^{k}-\alpha \tilde{\A}\nabla 	f(\ux^{k})-\tilde{\B}\uy^{k},\\
			\uy^{k+1}&=\uy^{k}+ \tilde{\B} \ux^{k},\\
		\end{aligned} 
	\end{equation}
	where $\tilde{\A} , \tilde{\C}  \in \R^{nd \times nd}$ are doubly stochastic matrices, $\alpha$ denotes the step-size, and $\y^k$ is the auxiliary variable used to mitigate errors and biases arising from the heterogeneity in local loss functions and decentralized processing. Futhermore, $\tilde{\B}  \in \R^{nd \times nd}$ imposes consensus. The online variant of above update rule~\eqref{ga} can be viewed as the special case of our GGT algorithm~\eqref{compact} when  $\H^{(4)} \H^{(2)} =\H^{(2)} \H^{(4)}$ and matrices in $\cH$ are defined such that $\tilde{\A} = \H^{(2)} \H^{(5)}$,
	$\tilde{\B}^2 = \I - \H^{(1)} - \H^{(4)} + \H^{(4)} \H^{(1)} - \H^{(2)} \H^{(3)}$, and $\tilde{\A} \tilde{\C} = \H^{(4)} \H^{(1)} - \H^{(2)} \H^{(3)}$. The regret rate analysis of the online GGT algorithm is therefore also applicable to the online variant of \eqref{ga}. It is remarked that \eqref{ga} is also a special case of the Unified Decentralized Algorithm (UDA) proposed for (static) decentralized optimization \cite{alghunaim2020decentralized}, if the matrices $\mathcal{A}$ and $\mathcal{B}$ of UDA satisfy $\mathcal{A}\mathcal{B}^2=\mathcal{B}^2\mathcal{A}$.
	
	\begin{table*}[ht] 
		\centering
		\caption{Possible choices of matrices $\H^{(\ell)}$ resulting in different variants of the proposed algorithm.}\label{newformtable}
		\def\arraystretch{1.5}
		\begin{tabular}{|c|c|c|c|c|c|c|c|c|}
			\hline 
			\diagbox{$\cH$}{Algo.}	&D-EXTRA  &D-NIDS &D-PGPDA  &D-GE &D-ATC-GT &D-GT &DOO-GT &oGGT\\
			\hline
			$\H^{(1)}$ &$\O_{nd}$ &$\O_{nd}$  &$\O_{nd}$  &$\W \otimes \I_d$ &$\W \otimes \I_d$  &$\W \otimes \I_d$   &$\W \otimes \I_d$ &$\W \otimes \I_d$\\
			$\H^{(2)}$ &$\I_{nd}$ &$\I_{nd}$   &$\I_{nd}$  &$\eta \I_{nd}$  &$\eta \W \otimes \I_d$ &$\W\otimes \I_d$ &$\eta \W \otimes \I_d$ &$\eta_1 \I_{nd} + \eta_2 \W\otimes \I_d$\\
			$\H^{(3)}$ &$-\V \otimes \I_d$ &$-\frac{1}{2}(\I+\W)\otimes \I_d$  &$-\frac{1}{2}(\I_{nd}+\tilde{\Z})$  &$\mathbf{0}_{nd}$ &$\mathbf{0}_{nd}$  &$\mathbf{0}_{nd}$ &$\mathbf{0}_{nd}$ &$\mathbf{0}_{nd}$\\
			$\H^{(4)}$ &$(\I+\W)\otimes \I_d$ &$(\I+\W)\otimes \I_d$   &$(\I_{nd}+\tilde{\Z})$  &$\W\otimes \I_d$ &$\W\otimes \I_d$ &$\W\otimes \I_d$  &$\W\otimes \I_d$ &$\W\otimes \I_d$ \\
			$\H^{(5)}$ &$\eta\I_{nd}$ &$\frac{1}{2}\eta(\I+\W)\otimes \I_d$
			&$\frac{\eta}{2} \tilde{\E}^{-1}$  &$\I_{nd}$  &$\W\otimes \I_d$ &$\eta \I_{nd}$ &$\I_{nd}$ &$\eta_3 \I_{nd} + \eta_4 \W\otimes \I_d$ \\
			$\H^{(6)}$ &$-\i^\T \otimes \I_d $ &$-\i^\T \otimes \I_d $
			&$-\i^\T \E \otimes \I_d$  &$\O_{d \times nd}$ &$\O_{d \times nd}$ &$\O_{d \times nd}$ &$\O_{d \times nd}$ &$\O_{d \times nd}$ \\
			$\H^{(7)}$ &$\i^\T \otimes \I_d $ &$\i^\T \otimes \I_d $
			&$\i^\T \E \otimes \I_d$  &$\i^\T \otimes \I_d $  &$\i^\T \otimes \I_d $  &$\i^\T \otimes \I_d $ &$\i^\T \otimes \I_d $ &$\i^\T \otimes \I_d $ \\
			$\H^{(8)}$ &$\eta \i^\T \otimes \I_d $ &$\eta \i^\T \otimes \I_d $
			&$\frac{\eta}{2}\i^\T \I_n \otimes \I_d$  &$\i^\T \otimes \I_d $  &$\i^\T \otimes \I_d $ &$\eta \i^\T \otimes \I_d $  &$\i^\T \otimes \I_d $ &$ (\eta_3 + \eta_4) \i^\T \otimes \I_d $ \\
			\hline
		\end{tabular}
	\end{table*}

	\section{Regret Rate Analysis}
	\label{secregret}
	
	In this section, we derive the regret performance of the proposed GGT algorithm by utilizing a state-space representation of the updates. We begin with stating the relevant assumptions and then proceed to establish some preliminary results that will be useful in characterizing the performance of GGT algorithm and its specializations. 
	
	\subsection{Assumptions} \label{subsec:assm}
	We begin with the following assumption that can be seen as a generalization of the strong convexity and smoothness assumptions standard in most decentralized online optimization algorithms. 
	\begin{assumption}\label{a1}
		\textit{$\M_i$-convexity and smoothness}: The functions $f_i^k:\Rn^d \rightarrow \Rn$ with  dom$(f) = \Rn^d$ are proper, closed, $L_i$-smooth, and $\M_i$-convex, i.e., $f_i^k(\x) - \frac{1}{2}\x^\T\M_i\x$ is convex. Such that, $\frac{1}{n}\sum_{i=1}^n \M_i > \mu \I,$ i.e., $f^k$ is $\mu-$strongly convex.
	\end{assumption}
	
	We note that for a $\mu$-strongly convex function $f_i^k$, we have that $\M_i = \mu\mathbf{I}_d$. An implication of Assumption \eqref{a1} is the following generalized co-coercivity property.
	
	\begin{lemma}\label{cor}
		If a function $f_i^k$ satisfies Assumption $\ref{a1}$ such that $L_i\I_d - \M_i$ is invertible, then it holds that
		\begin{align}
			\begin{bmatrix} \x-\y\\ \u - \v \end{bmatrix}^{\T} \S_i \begin{bmatrix} \x-\y\\ \u - 	\v \end{bmatrix}\geq 0. \label{coco}
		\end{align}
		Where $\u = \nabla f_i^k(\x)$, $\v = \nabla f_i^k(\y)$, 
		\begin{align}
			\S_i \coloneqq \begin{bmatrix} -2 L_i \L_{\M_i}^{-1}\M_i & \L_{\M_i}^{-1}(\M_i +L_i 	\I)\\ \L_{\M_i}^{-1}(\M_i +L_i \I) & -2\L_{\M_i}^{-1} \end{bmatrix}, \label{scosm}
		\end{align}
		and $\L_{\M_i} = L_i \I_d - \M_i$, for all $\x$, $\y \in \Rn^d$. 
	\end{lemma}
	The proof of Lemma \ref{cor} is provided in the Supplementary material and follows along the lines of the proof of co-coercivity. From \eqref{fav}, the overall function $f^k$ is also $L$-smooth and $\mu$-strongly convex for $L = \frac{1}{n}\sum_{i=1}^n L_i$ and $\mu \I > \frac{1}{n}\sum_{i=1}^n \M_i$. Assumption \eqref{a1} also implies that the global function gradient vanishes at $\x^{k,\star}$, i.e.,$\nabla f^k(\x^{k,\star}) = 0$. 
	The next two assumptions relate to the dynamics of the optimum $\x^{k,\star}$. 
	\begin{assumption}
		\label{a2}
		\textit{Bounded variations}: the norm of the difference between consecutive optima is bounded as $\norm{\xks - \x^{(k-1),\star}} \leq \sigma_1$ for $k \geq 2$, where $\sigma_1 \geq 0$. 
	\end{assumption}
	\begin{assumption}
		\label{a3}
		\textit{Bounded gradient variations}: the norm of the difference between consecutive gradients of local loss functions evaluated at the respective optima is bounded as $\norm{\nabla f_i^k(\xks) - \nabla f_i^{k-1}(\x^{(k-1),\star} )} \leq \sigma_2$ for all $k \geq 2$, where $\sigma_2 \geq 0$.
	\end{assumption}
	
	Assumptions \eqref{a2}-\eqref{a3} impose restrictions on the sudden changes in $\xks$ or in $\nabla f_i^k(\xks)$ across $k$. Observe that these bounds do not follow from the sublinearity of $C_K$ and $D_K$, which characterize the growth in the cumulative path lengths and cumulative gradient differences, respectively. Instead, Assumptions \eqref{a2}-\eqref{a3} only impose bounds on the individual summands in \eqref{ck}-\eqref{dk}. 
	
	\subsection{Compact Form}\label{subsec:compact_form}
	The updates in \eqref{compact} and \eqref{initialz_cmpct0} can be compactly written as 
	\begin{subequations}\label{uniupdate}
		\begin{align}
			\z^{k+1} &= \A\z^k+\B\u^k, \label{akbk} \\ 
			\y^k &= \C\z^k+\D\u^k, 	\label{ckdk} \\ 
			\O &= \F \z^k + \G \u^k, \label{fkgk}
		\end{align}
	\end{subequations}
	where the state $\z^k$ and the gradient $\u^k$ are defined as
	\begin{align}\label{zdef}
		\z^k &= \begin{bmatrix}
			\ux^{k-1} \\
			\us^{k-1}  \\
			\nab^{k-1}
		\end{bmatrix}, & \u^k &= \nabla f^k(\y^k), 
	\end{align}
	and
	\begin{align*}
		\A &=\begin{bmatrix} \H^{(1)} &\H^{(2)} &\O \\
			\H^{(3)} &\H^{(4)} &\H^{(5)} \\
			\O &\O &\O
		\end{bmatrix},
		\quad \B=\begin{bmatrix}
			\O \\
			-\H^{(5)} \\
			\I
		\end{bmatrix},\\
		\C &=\begin{bmatrix} \H^{(1)} &\H^{(2)} &\O
		\end{bmatrix},
		\qquad \quad  \quad \D=\begin{bmatrix}
			\O
		\end{bmatrix},\\
		\F &=\begin{bmatrix} \H^{(6)} &\H^{(7)} &\H^{(8)}
		\end{bmatrix},
		\qquad \G= \begin{bmatrix}
			\O
		\end{bmatrix}.
	\end{align*}
	Let $\{\z^{k,\star}, \u^{k,\star}\}$ be the fixed point of \eqref{uniupdate}, so that 
	\begin{subequations}\label{uniupdateopt}
		\begin{align}	
			\z^{k,\star} &= \A\z^{k,\star}+\B\u^{k,\star} \label{akbkstar}, \\
			\y^{k,\star} &= \C\z^{k,\star}+\D\u^{k,\star} \label{ckdkstar}, \\
			\O &= \F \z^{k,\star} + \G \u^{k,\star}. \label{fkgkstar}
		\end{align}
	\end{subequations}
	The above form allows us to construct an SDP that captures the regret rate of the proposed algorithm. 
	
	\subsection{Preliminary Results}
	\label{subsec:premres}
	The following Theorem provides the key preliminary result that relates the distance between an iterate and the optimal value at that iteration using an SDP. 
	\begin{theorem} \label{mytheo1} 
		Let $\R$ be a real matrix whose columns are the basis vectors of the null space of $[\F \quad \G]$. Under Assumption \eqref{a1}, if there exist $\rho > 0$, $\P \succ 0$, and $\lambda \geq 0$ such that 
		\begin{align} 
			\label{SDP}
			\R^\T&\left(\begin{bmatrix} \A^{\T}\P\A-\rho^{2}\P & \A^{\T}\P\B\\ \B^{\T}\P\A 	& 	\B^{\T}\P\B \end{bmatrix}+\lambda \begin{bmatrix} \C & \D\\ \O & \I \end{bmatrix}^{\T}\M\begin{bmatrix} \C & \D\\ \O & \I \end{bmatrix}\right)\R\preceq 0,
		\end{align}
		where, $\M$ is defined as below
		$$ \begin{bmatrix} \text{Diag}(\{-2L_i \L_{\M_i}^{-1} \M_i\}_{i=1}^n) & \text{Diag}( \{\L_{\M_i}^{-1}(\M_i+ L_i \I_d)\}_{i=1}^{n})  \\
			\text{Diag}( \{\L_{\M_i}^{-1}(\M_i+ L_i \I_d)\}_{i=1}^{n})  &-2 \text{Diag}( 	\{\L_{\M_i}^{-1}\}_{i=1}^{n})
		\end{bmatrix},$$
		$\L_{\M_i} = L_i \I_d - \M_i $
		then $\Vert{\z^{k+1}-\z^{k,\star}}\Vert_\P \le \rho \Vert{\z^k-\z^{k,\star}}\Vert_\P$ for all $k \ge 0$.  
	\end{theorem}

	\begin{IEEEproof}
		Subtracting \eqref{fkgk} from \eqref{fkgkstar}, we observe that $[(\z^k-\z^{k,\star})^{\T} \quad (\u^k-\u^{k,\star})^{\T}]$ lies in the null space of $[\F \quad \G]$. Therefore, it must hold that $\R\v = [(\z^k-\z^{k,\star})^{\T} \quad (\u^k-\u^{k,\star})^{\T}]^\T$ for some $\v$. Since the matrix on the left of \eqref{SDP} is negative-semidefinite, it follows that 
		\begin{align} 
			&[(\z^k-\z^{k,\star})^{\T} (\u^k-\u^{k,\star})^{\T}] \left(\begin{bmatrix} 	\A^{\T}\P\A-\rho^{2}\P & \A^{\T}\P\B\\ \B^{\T}\P\A & \B^{\T}\P\B \end{bmatrix} +\lambda \begin{bmatrix} \C & \D\\ \O & \I \end{bmatrix}^{\T}\M \begin{bmatrix} \C & \D \\ \O & \I \end{bmatrix}\right) \begin{bmatrix} \z^k-\z^{k,\star} \\  \u^k-\u^{k,\star} \end{bmatrix} \leq 0. \label{refSDP} 
		\end{align}
		Now, after simplifying the first term of the above equation, we have
		\begin{align}
			&\begin{bmatrix} \z^k-\z^{k,\star} \\  \u^k-\u^{k,\star} \end{bmatrix}^\mathsf{T} 	\begin{bmatrix} \A^{\T}\P\A-\rho^{2}\P & \A^{\T}\P\B\\ \B^{\T}\P\A & \B^{\T}\P\B \end{bmatrix}\begin{bmatrix} \z^k-\z^{k,\star} \\  \u^k-\u^{k,\star} \end{bmatrix} =(\z^k-\z^{k,\star})^{\T}\A^{\T}\P(\A\z^k-\A\z^{k,\star}+\B\u^k-\B\u^{k,\star}) 	\nonumber \\
			&~~~~~~~~~~~~~~~~~~~~~~~+(\u^k-\u^{k,\star})^{\T}\B^{\T}\P(\A\z^k-\A\z^{k,\star}+ \B\u^k-\B\u^{k,\star})-\rho^2(\z^k-\z^{k,\star})^{\T}\P(\z^k-\z^{k,\star}). \label{sdpproof1}
		\end{align}
		Here, the first two terms can be simplified through the repeated use of \eqref{akbk} and \eqref{akbkstar} as follows:
		\begin{align}
			(\z^k-\z^{k,\star})^{\T}&\A^{\T}\P(\A\z^k-\A\z^{k,\star}+\B\u^k-\B\u^{k,\star}) 	+(\u^k-\u^{k,\star})^{\T}\B^{\T}\P(\A\z^k-\A\z^{k,\star}+ \B\u^k-\B\u^{k,\star}) 	\nonumber\\
			&= (\z^{k+1}-\z^{k,\star})^\T \P (\A\z^k-\A\z^{k,\star}+\B\u^k-\B\u^{k,\star}) 	\nonumber\\
			&= (\z^{k+1}-\z^{k,\star})^{\T}\P (\z^{k+1}-\z^{k,\star}).\label{sdpproof2}
		\end{align}
		Substituting \eqref{sdpproof2} into \eqref{sdpproof1} and using the definition of $\P$-norm, we obtain
		\begin{align}
			&\begin{bmatrix} \z^k-\z^{k,\star} \\  \u^k-\u^{k,\star} \end{bmatrix}^\mathsf{T} 	\begin{bmatrix} \A^{\T}\P\A-\rho^{2}\P & \A^{\T}\P\B\\ \B^{\T}\P\A & \B^{\T}\P\B \end{bmatrix}\begin{bmatrix} \z^k-\z^{k,\star} \\  \u^k-\u^{k,\star} \end{bmatrix} = \norm{\z^{k+1}-\z^{k,\star}}_{\P}^2-\rho^2 \norm{\z^{k}-\z^{k,\star}}_{\P}^2. \label{SDP_1_term}
		\end{align}  
		The second term of \eqref{refSDP} can be simplified along similar lines by using \eqref{ckdk} and \eqref{ckdkstar}, so as to yield
		\begin{align} 
			&\lambda \begin{bmatrix} \z^k-\z^{k,\star} \\  \u^k-\u^{k,\star} 	\end{bmatrix}^\mathsf{T} \begin{bmatrix} \C & \D\\ \O & \I \end{bmatrix}^{\T}\M \begin{bmatrix} \C & \D\\ \O & \I \end{bmatrix} \begin{bmatrix} \z^k-\z^{k,\star} \\  \u^k-\u^{k,\star} \end{bmatrix} = \lambda \begin{bmatrix} \y^k-\y^{k,\star}\\ \u^{ k}-\u^{k,\star} 	\end{bmatrix}^{\T}\M\begin{bmatrix} \y^k-\y^{k,\star}\\ \u^{ k}-\u^{k,\star} \end{bmatrix} \gfrom{coco} 0. \label{SDP_2_term}
		\end{align}
		The non-negativity of the second term in \eqref{refSDP} implies that it can be dropped from the left-hand side, so as to obtain
		\begin{align}
			\norm{\z^{k+1}-\z^{k,\star}}_{\P}^2  &\leq \rho^2 \norm{\z^{k}-\z^{k,\star}}_{\P}^2. 	\nonumber
		\end{align}
		Upon taking square root on both sides, we obtain the required result. 
	\end{IEEEproof}
	
	Theorem \eqref{mytheo1} establishes that if the linear matrix inequality (LMI) in \eqref{SDP} has a feasible solution for some $\rho<1$, the update in \eqref{uniupdate} brings the state closer to its fixed point. It is remarked that for a given algorithm, the sufficient condition in Theorem \ref{mytheo1} depends on the problem parameters $\{\M_i, L_i\}$ and on the parameters $\rho$, $\P$ and $\lambda$, which have to be found using numerical search, as detailed later in Sec. \ref{tuning}. If no such parameters exist, then the subsequent results are also not applicable. In particular, under Assumption \eqref{a1}, Lemma \ref{cor}, and if \eqref{SDP} is satisfied for some $\rho \in (0,1)$, the distance from the optimum after the update at time $k$, given by $\norm{\z^{k+1} - \z^{k,\star}}_\P$, is at most a fraction of the original distance from the optimum, given by $\norm{\z^k - \z^{k,\star}}_\P$. Theorem \ref{mytheo1} will subsequently be used to obtain the conditions under which the proposed GGT algorithm achieves a sublinear dynamic regret. A similar SDP-based result is available for the static and $\mu$-strongly convex case in \cite[Lemma 2]{sundararajan2017robust}. The construction in Theorem \ref{mytheo1} is different as it generalizes the existing results and pertains to the dynamic case, where the optimum values are also time-varying. We further observe that although $\M \in \Rn^{2nd \times 2nd}$, the size of the SDP can be reduced for the case when $f_i^k$ are $\mu$-strongly convex by writing $\M$ as a Kronecker product of $\I_d$ with an $2n \times 2n$ matrix, and performing further simplifications in \eqref{SDP}. 
	
	The specific form of $\z^k$ and $\z^{k,\star}$ is algorithm-dependent, and is provided in Table \ref{defznzs} for each of the algorithms considered here. The statement of Theorem \ref{mytheo1} may also be applicable to other algorithms whose updates can be written as in \eqref{uniupdate} and whose optimum satisfies \eqref{uniupdateopt}. We remark that for $\z^k$ defined as in \eqref{zdef}, the bounds in Assumptions \eqref{a2}-\eqref{a3} translate to a bound of the form 
	\begin{align}
		\norm{\z^{k,\star}-\z^{k-1,\star}}_{\P} \leq \sigma, \label{zbound}
	\end{align}
	where $\sigma \geq 0$ depends on $\P$, $\sigma_1$, and $\sigma_2$. Intuitively, \eqref{zbound} says that since sudden changes in $\x^{k,\star}$ and $\nabla f_i^k(\x^{k,\star})$ are bounded, the variations in the super-vector $\z^{k,\star}$ are also bounded. Such a bound is necessary, because the unified proof of GGT is developed in terms of $\z^{k,\star}$.
	
	An implication of \eqref{zbound} is that for $\rho \in (0,1)$ the result in Theorem~\ref{mytheo1} ensures that the iterates generated by the GGT algorithm stay within a bounded distance from the optimum for all time $k \geq 1$. The following corollary, whose proof is provided in Appendix~\ref{col_pr: zk1_zs_bnd}, gives the precise statement of this bound. 
	\begin{corollary} \label{col: zk1_zs_bnd}
		Under Assumptions~\ref{a1}-\ref{a3} and if there exist $\rho \in (0,1)$  and $\P \succ 0$	such that Theorem \ref{mytheo1} holds, then we have that
		\begin{align}
			\Vert \z^{k+1}-\z^{k,\star}\Vert \le & \sqrt{\text{cond}(\P)}\rho^k\Vert 		\z^1-\z^{1,\star}  \Vert +\frac{\sigma\rho(1-\rho^{k-1})}{(1-\rho)(\sqrt{\lambda_{\min}(\P)})}.\label{col1eq}
		\end{align}
	\end{corollary}
	The proof of Corollary \ref{col: zk1_zs_bnd} is provided in Appendix \ref{col_pr: zk1_zs_bnd} and follows from expanding the left-hand side of \eqref{col1eq} through recursive application of the result of Theorem \ref{mytheo1} and the use of \eqref{zbound}. 
	
	The importance of Corollary \ref{col: zk1_zs_bnd} is highlighted by the following bound on gradient norm. 
	\begin{corollary}\label{col:grad_bnd}
		Under Assumptions~\ref{a1}-\ref{a3} and if there exist $\rho \in (0,1)$ and $\P \succ 0$	such that Theorem \ref{mytheo1} holds, then the gradient norm is bounded as
		\begin{align*}
			&\Vert \nabla f^k(\x_i^k) \Vert \leq \frac{L \rho}{\sqrt{\lambda_{\min}(\P)}}\max\{\frac{\sigma 	(1-\rho^{k-1})}{(1-\rho)},\rho^{k-1} \Vert \z^1-\z^{1,\star} \Vert_{\P}\}.
		\end{align*}
	\end{corollary}
	The proof of Corollary~\ref{col:grad_bnd} is provided in Appendix~\ref{col_pr:grad_bnd}. The bound in Corollary~\ref{col:grad_bnd} is unique and different from the much stronger bounded gradient assumption that is commonly made in the literature; see \cite{shahrampour2018distributed,nazari2021adaptive,wu2022decentralized,zhang2019distributed}. The bound in Corollary~\ref{col:grad_bnd} is intuitively satisfying since it is 	applicable to the global function gradient, which is zero at $\x^{k,\star}$. In contrast, an assumption on the boundedness of $\norm{\nabla f_i^k(\x)}$ would be significantly stronger, since the local function gradients are not necessarily zero at $\x^{k,\star}$. 
	
	In a similar vein, we define the track length of $\z^{k,\star}$ as
	\begin{equation}
		\label{pathlength}
		W_{K,p} \coloneqq \sum_{k=2}^{K} \Vert \z^{k, \star}-\z^{(k-1), \star} \Vert_\P^p,
	\end{equation}
	where $p\geq 1$. The subsequent analysis will yield bounds that depend on both $W_{K,1}$ (represented by $W_K$) and $W_{K,2}$. 
	In the specific case where the definition of $\z^k$ \eqref{zdef} and $\z^{k,\star}$, is of the form as listed in Table~\ref{defznzs} we have the following result whose proof is provided in Appendix~\ref{lem_pr:wk_ck_dk}.
	\begin{lemma} \label{lem:wk_ck_dk}
		The track length $W_K$ is bounded as 
		\begin{align}
			W_K &\leq \sqrt{2\lambda_{\max}(\P)} C_K +\sqrt{2\lambda_{\max}(\P)} D_K. 	\label{eq:lem:wk_ck_dk} \\
			W_{K,2} & \leq 2 \lambda_{\max}(\P) C_{K,2} + 2 \lambda_{\max}(\P) D_{K,2}. 	\label{eq:lem:wk2_ck2_dk2}
		\end{align}
	\end{lemma}
	Lemma \ref{lem:wk_ck_dk} establishes that $W_K = \cO(C_K+D_K)$, $W_{K,2} = \cO(C_{K,2} + D_{K,2} )$ and will be used for specializing the bound obtained for GGT to its variants listed in Table \ref{newformtable}.

	\begin{table}[ht] 
		\centering
		\caption{Definitions of $\{\z^k,\u^k,\z^{k,\star}, \u^{k,\star}\}$ for different algorithms}\label{defznzs}
		\begin{center}
			\def\arraystretch{1.5}
			\begin{tabular}{|c|c|c|c|c|}
				\hline 
				\diagbox{Algo.}{Vect.}	&$\z^k$  &$\u^k$ &$\z^{k,\star}$ &$\u^{k,\star}$ \\
				\hline
				\begin{tabular}[c]{@{}c@{}}D-EXTRA,\\ D-NIDS, \\D-PGPDA \end{tabular}
				&$\begin{bmatrix}
					\ux^{k-1} \\
					\ux^k  \\
					\nab^{k-1}
				\end{bmatrix}$ &$\nabla f^k(\y^k)$ &$\begin{bmatrix}
					\ux^{k,\star} \\
					\ux^{k,\star}  \\
					\nab^{k,\star}
				\end{bmatrix}$ &$\nabla f^{k}(\y^{k,\star})$\\ \hline
				\begin{tabular}[c]{@{}c@{}}D-GE, D-GT, \\D-ATC-GT, \\DOOGT, \\oGGT \end{tabular} &$\begin{bmatrix}
					\ux^{k-1} \\
					\ug^{k-1}  \\
					\nab^{k-1}
				\end{bmatrix}$ &$\nabla f^k(\y^k)$ &$\begin{bmatrix}
					\ux^{k,\star} \\
					\O  \\
					\nab^{k,\star}
				\end{bmatrix}$ &$\nabla f^{k}(\y^{k,\star})$\\
				\hline
			\end{tabular}
		\end{center}
	\end{table}
	
	\subsection{Regret Analysis} \label{susec:reg_analy}
	We begin with following preliminary result, which bounds the distance from the optimum in terms of the track length. 
	\begin{lemma}\label{lem2}
		If there exist $\rho \in (0,1)$ and $\P \succ 0$ that satisfy \eqref{SDP}, then the GGT iterate error and the sum of square of the GGT iterate error are bounded as
		\begin{equation}
			\label{linear bound}
			\sum _{k=1}^{K} \Vert \z^k-\z^{k, \star} \Vert_\P \leq \frac{\Vert \z^1 - \z^{1, \star} \Vert_\P + W_K}{1-\rho},
		\end{equation}
		\begin{equation}
			\label{quad_bound}
			\sum_{k=1}^K  \norm{\z^{k} - \z^{k,\star} }_{\P}^2 	\leq \frac{2\norm{\z^{1} - \z^{1,\star} }_{\P}^2 + 4 W_{K,2} } {(1-\rho^2)^2}. \\
		\end{equation}
	\end{lemma}
	The proof of Lemma \ref{lem2} is provided in the Appendix \ref{pr_lem2} and proceeds by expanding the sum in \eqref{linear bound} and \eqref{quad_bound}, using \eqref{SDP}, and simplifying the result utilizing Young's inequality.

	Having obtained the bound on the GGT iterate error, we are ready to state the main regret bound of the GGT algorithm. We have derived the bound for the general case and for the special case under Assumptions~\ref{a2} and \ref{a3}. 
	
	The next theorem bounds the dynamic regret of the GGT algorithm in the general case.
	\begin{theorem} \label{mytheo3}
		Under Assumption~\eqref{a1} if there exist $\rho \in (0, 1) $ and $\P\succ 0$ such that Theorem \ref{mytheo1} holds, then the dynamic regret of the GGT algorithm is bounded as:
		\begin{align}
			\Reg^K &\leq \frac{2L \rho^2}{n \lambda_{\min}(\P) (1-\rho^2)^2} \Bigg( 	\norm{\z^{1}-\z^{1,\star} }_{\P}^2 + W_{K,2} \Bigg). \nonumber 
		\end{align}
	\end{theorem}
	\begin{proof}
		We begin with utilizing the smoothness property of global function $f^k$, as $f^k$ is $L$-smooth, we have, 
		\begin{align}
			\Reg^K &= \tfrac{1}{n}\sum_{j=1}^n \sum_{k=1}^K \left(f^k(\x_j^k)-f^k(\xks)\right), 	\nonumber \\
			&\leq \frac{L}{2n}\sum_{j=1}^n \sum_{k=1}^K  \norm{\x_j^k - \xks}^2  = \frac{L}{2n} 	\sum_{k=1}^K  \norm{\underline{\x}^k - \underline{\x}^{k,\star}}^2. \nonumber
		\end{align}
		As $\ux^k$ and $\ux^{k,\star}$ are corresponding components of $\z^{k+1}$ and $\z^{k,\star}$ respectively, we can write
		\begin{align}
			\Reg^K &\leq \frac{L}{2n \lambda_{\min}(\P)} \sum_{k=1}^K  \norm{\z^{k+1} - 	\z^{k,\star} }^2_{\P}. \label{eq:lem:reg_def_sqr}
		\end{align}
		We can further simplify \eqref{eq:lem:reg_def_sqr} using the result of Theorem~\ref{mytheo1} and \eqref{quad_bound} to obtain the final result. 
	\end{proof}
	By combining Theorem~\ref{mytheo3} with Lemma \ref{lem:wk_ck_dk}, we derive a regret bound expressed in terms of both the cumulative squared path length and the cumulative squared gradient difference. 
	\begin{corollary} \label{col: main_result_sqr}
		Under Assumptions~\eqref{a1} and if there exist $\rho \in (0,1)$ and $\P\succ 0$ such that Theorem \ref{mytheo1} holds, the GGT algorithm variants, namely D-EXTRA, D-NIDS, DDLM, D-PGPDA, D-GE, D-GT and oGGT achieve a dynamic regret of $\cO(1+C_{K,2}+D_{K,2})$. 
	\end{corollary}
	The proof for Corollary \ref{col: main_result_sqr} is directly obtained by inserting the outcome of Lemma \ref{lem:wk_ck_dk} into the regret bound defined in Theorem \ref{mytheo3}. It is worth noting that Corollary \ref{col: main_result_sqr} can also be applied to any GGT variation that adheres to updates consistent with \eqref{uniupdate}-\eqref{uniupdateopt}.
	
	In some cases, the individual summands $\norm{\z^{k,\star}-\z^{k-1,\star}}^2$ may be very large, even if total track-length $W_{K,2}$ is sublinear. We can obtain tighter regret bounds for these cases under Assumptions~\eqref{a2}-\eqref{a3} as stated in the subsequent theorem.
	
	\begin{theorem} \label{mytheo2}
		Under Assumptions~\eqref{a1}-\eqref{a3} and if there exist $\rho \in (0,1)$ and $\P\succ 0$ such that Theorem \ref{mytheo1} holds, then the dynamic regret of the GGT algorithm is bounded as:
		\begin{align}
			\Reg^K &\leq \frac{L \sigma \rho^2}{\lambda_{\min}(\P)(1-\rho)^2}(\Vert \z^1 - 	\z^{1, \star} \Vert_\P + W_K).\label{finalregret}
		\end{align}
	\end{theorem}
	\begin{IEEEproof}
		We begin with utilizing the first order convexity property of global function $f^k$, as $f^k$ is convex, we have, 
		\begin{equation*}
			f^k(\x_j^k)-f^k(\xks) \leq (\nabla f^k(\x_j^k))^\T(\x_j^k-\xks).
		\end{equation*}
		From Cauchy-Schwartz inequality and Corollary~\ref{col:grad_bnd} we have
		\begin{align}
			f^k(\x_j^k)-f^k(\xks) &\leq \Vert \nabla f^k(\x_j^k) \Vert \Vert (\x_j^k-\xks)\Vert 	\nonumber \\
			&\leq G \Vert (\x_j^k-\xks)\Vert, \label{theo_pr_ref1}
		\end{align}
		where $G=  \frac{L \rho}{\sqrt{\lambda_{\min}(\P)}} \max\{\sigma_3,\rho^{k-1} \Vert \z^1-\z^{1,\star} \Vert_{\P}\}$ and $\sigma_3 = \frac{\sigma (1-\rho^{k-1})}{(1-\rho)}. $ On substituting the bound present in \eqref{theo_pr_ref1}, in dynamic regret definition \eqref{dyreg} we obtain,
		\begin{align}
			\Reg^K & = \frac{1}{n}\sum_{j=1}^n \bigg{(}\sum_{k=1}^K (f^k(\x_j^k)- 		f^k(\xks))\bigg{)} \nonumber \\
			&\leq  \frac{1}{n}\sum_{j=1}^n \bigg{(}\sum_{k=1}^K  G \Vert (\x_j^k-\xks)\Vert 	\bigg{)}. \label{theo_pr_ref2}
		\end{align}
		From the property of norm we have $ \left( \sum_{j=1}^{n}\Vert (\x_j^k-\xks)\Vert \right)^2 \leq n\norm{\ux^k-\ux^{k,\star}}^2$. On taking square root on both the sides we get $\sum_{j=1}^{n}\Vert (\x_j^k-\xks)\Vert \leq \sqrt{n}\norm{\ux^k-\ux^{k,\star}}$, which on substituting in \eqref{theo_pr_ref2} we get
		\begin{align}
			\Reg^K &= \frac{G}{\sqrt{n}}\bigg{(}\sum_{k=1}^K  \Vert (\ux^k-\ux^{k,\star})\Vert 	\bigg{)}. \nonumber
		\end{align}
		As $\ux^k$ and $\ux^{k,\star}$ are corresponding components of $\z^{k+1}$ and $\z^{k,\star}$, so we have
		\begin{align}
			\Reg^K &\leq \frac{G}{\sqrt{n}}\bigg{(}\sum_{k=1}^K  \Vert 	(\z^{k+1}-\z^{k,\star})\Vert \bigg{)} \label{theo_pr_ref3}.
		\end{align}
		On applying Theorem~\ref{mytheo1} we can further simplify \eqref{theo_pr_ref3} as below
		\begin{align}
			\Reg^K &\leq \frac{G}{\sqrt{n}}\bigg{(}\sum_{k=1}^K  \rho \Vert 	(\z^k-\z^{k,\star})\Vert \bigg{)}, \nonumber\\
			&\leq \frac{G \rho}{\sqrt{n \lambda_{\min}(\P)}} \sum_{k=1}^K \Vert 	(\z^k-\z^{k,\star})\Vert_{\P}. \nonumber
		\end{align}
		We obtain the desired result by substituting the result from Lemma~\ref{lem2}.
	\end{IEEEproof}
	On combining Theorem~\ref{mytheo2} and Lemma \ref{lem:wk_ck_dk}, we get the following regret bound in terms of the cumulative path length and the cumulative gradient difference. 
	\begin{corollary} \label{col: main_result}
		Under Assumptions~\eqref{a1}-\eqref{a3} and if there exist $\rho > 0$ and $\P\succ 0$ such that Theorem \ref{mytheo1} holds, the GGT algorithm variants, namely D-EXTRA, D-NIDS, DDLM, D-PGPDA, D-GE, and D-GT, achieve a dynamic regret of $\cO(1+C_K+D_K)$. 
	\end{corollary}
	
	The proof of Corollary \ref{col: main_result} follows immediately from substituting the result of Lemma \ref{lem_pr:wk_ck_dk} into the regret bound in Theorem \ref{mytheo2}. Combining Corollaries~\ref{col: main_result_sqr} and \ref{col: main_result},  we derive the final dynamic regret bound.
	\begin{corollary} \label{col: main_result_final}
		Under Assumptions~\eqref{a1}-\eqref{a3} and if there exist $\rho \in (0,1)$ and $\P\succ 0$ such that Theorem \ref{mytheo1} holds, the GGT algorithm variants, namely D-EXTRA, D-NIDS, DDLM, D-PGPDA, D-GE, and D-GT, achieve a dynamic regret of $\cO(1+ \min \{ C_K+D_K , C_{K,2} + D_{K,2}\})$. 
	\end{corollary}
	We remark that $C_{K,2} + D_{K,2}$ may be smaller than $C_K + D_K$ if all the consecutive variations are small. Conversely, if even one of $\norm{\x^{k,\star}-\x^{k-1,\star}}$  or $\norm{\boldsymbol{\nabla}^{k,\star}-\boldsymbol{\nabla}^{k-1,\star}}$ is large for any $k$, then $C_K+D_K$ would be smaller than $C_{K,2}+D_{K,2}$. Further, in the second case, when the individual variations can be large, Assumptions \ref{a2} and \ref{a3} are needed to prevent them from being unbounded. Similar assumptions have appeared in earlier works \cite{dixit2019online, derenick2009optimal, derenick2009convex}.
	
	The dynamic regret bound in Corollary \ref{col: main_result_final} is contingent on Theorem \ref{mytheo1} holding for some $\rho \in (0,1)$, $\P \succ \mathbf{0}$ and $\lambda \geq 0$. Theorem 1 can hence be seen as a numerical approach for checking the applicability of various dynamic regret bounds obtained in this work. Indeed, it is not necessary that the parameters in Theorem \ref{mytheo1} can always be found. For instance, we can write the DOG algorithm as a special case of GGT with $\H^{(1)} = \O, \quad  \H^{(2)} = \I, \quad  \H^{(3)} = -\W, \quad \H^{(4)} = \W + \I, \quad \H^{(5)} = \eta \I \quad \H^{(6)}=\H^{(7)} = \H^{(8)} = \mathbf{0}$, and 
	\begin{align*}
		\z^k &= \begin{bmatrix}
			\ux^{k-1} \\
			\ux^k  \\
			\nab^{k-1}     
		\end{bmatrix},  \qquad 
		\z^{k,\star} = \begin{bmatrix}
			\ux^{k, \star} \\
			\ux^{k, \star}  \\
			\nab^{k, \star}
		\end{bmatrix}.
	\end{align*}
	With this choice of parameters, while DOG does satisfy \eqref{uniupdate}-\eqref{uniupdateopt}, it turns out that the SDP in \eqref{SDP} is infeasible for all $\rho \in (0,1)$. Hence, Theorem \ref{mytheo1} cannot be used to obtain the dynamic regret of DOG. It is also known from the literature that the dynamic regret bound of DOG is strictly worse than that of GGT. 
	
	\subsection{Optimized GGT}\label{tuning}
	All algorithms in the literature, as well as the proposed GGT algorithm have several hyperparameters that must generally be tuned to obtain the best possible performance. In various machine learning applications, hyperparameters are generally tuned experimentally by optimizing the performance of the algorithm on a small subset of data. However, such an approach is not necessarily practical in distributed or online settings, where the data points arrive sequentially. In online settings therefore, it is desirable that the algorithm hyperparameters can be calculated a priori, using minimal information about the problem, such as from the network structure, mixing matrix $\W$, and the objective function parameters $\{L_i, \M_i\}$.

	For the different GGT variants proposed in Sec. \ref{secunif}, the step-size $\eta$ is the only hyperparameter that requires tuning. For each variant and given $(\W, \{L_i, \M_i\})$, Theorem \ref{mytheo1} allows us to check if a specific value of $\rho$ and $\eta$ returns a feasible LMI in \eqref{SDP}. Hence, the optimal value of $\eta$ for each algorithm corresponds to the one that minimizes the bound on the dynamic regret in  \eqref{finalregret}, while ensuring that the LMI in \eqref{SDP} is still feasible. In Appendix~\ref{pr_cond_P_rho}, we have shown that the regret depends on the expression $\frac{\rho^2}{ (1-\rho)^2} \text{cond}(\P)$. To minimize the regret, we seek the value of $\eta$ for which $\frac{\rho^2}{ (1-\rho)^2} \text{cond}(\P)$ is minimized, while ensuring that the SDP \eqref{SDP} is feasible.  Algorithm~\ref{HPtun} provides additional information regarding the procedure for finding the optimal hyperparameters of the GGT algorithm.
	\begin{algorithm}
		\caption{Pseudo code for hyperparameter tuning}\label{HPtun}
		\begin{algorithmic}[h]
			\State	\textbf{Input} $\A,\B,\C,\D,\F,\G,\M,\epsilon$ (tolerance)
			\State \textbf{Initialize} $\eta=\eta_0, \rho_l = 0, \rho_u = 1, C_{\min} = C_0, \text{FEAS} =0$
			\While{$\eta < \frac{2}{L}$} \qquad \qquad \qquad \qquad (Line-search)
			\While{$\abs{\rho_u - \rho_l} > \epsilon$} \qquad \qquad (Bisection)
			\State $\rho = \frac{\rho_l + \rho_u}{2}$, Solve the following problem
			\begin{align}
				\underset{\P, \lambda}{\min} \quad&\text{cond}(\P) \nonumber \\
				s.t.\quad &\R^\T\left(\begin{bmatrix} \A^{\T}\P\A-\rho^{2}\P & \A^{\T}\P\B\\ 		\B^{\T}\P\A & \B^{\T}\P\B \end{bmatrix} 
				+\lambda \begin{bmatrix} \C & \D\\ \O & \I 	\end{bmatrix}^{\T}\M\begin{bmatrix} 	\C & \D\\ \O & \I \end{bmatrix}\right)\R\preceq 0  \nonumber \\  
				& \lambda \geq 0,  P \succ 0. \nonumber
			\end{align} 
			\If{There exist a $\P, \lambda$}
			\State $\rho_u = \rho, C= \frac{\rho^2}{ (1- \rho )^2} \text{cond}(\P) $
			\If{$C < C_{\min}$}
			\State $\eta_{\text{opt}} = \eta, C_{\min} = C, \text{FEAS} = 1 $ 
			\EndIf
			\Else
			\State $\rho_l = \rho$
			\EndIf
			\EndWhile
			\State $\eta = \eta + 10\epsilon$
			\EndWhile
			\If{$\text{FEAS}=0$}
			\textbf{ Output} Infeasible
			\Else
			\textbf{ Output} $\eta_{opt}$
			\EndIf	
		\end{algorithmic}
	\end{algorithm} 
	\text{}\\
	For initialization, choose very small value of $\eta_0$, for example $\eta_0 = 0.01/L$, and a very large value of $C_{\min}$, for example $C_{\min} = 10^6$, $\epsilon$ is a tolerance for bisection, and we have also utilized it as a step-size for the line search of $\eta$.
	
	On the other extreme, one can consider the entire set of GGT matrices $\{\H^\ell\}_{\ell=1}^8$ as a hyperparameter block and attempt to tune it with the goal of minimizing the regret. However, such an endeavor might be intractable, given the large number of hyperparameters and associated constraints such as \eqref{initialz_cmpct0}-\eqref{Mequalities}. Instead, we consider a simplified version of GGT, where the mixing matrix $\W$ is specified according to the network, and the constituents of $\cH$ are forced to follow the template of D-GE, D-GT, and D-ATC-GT algorithms, as depicted in Table \ref{defznzs}. The resulting oGGT algorithm, whose parameter matrices are listed in the last column of Table \ref{newformtable}, therefore has only four free parameters $\{\eta_i\}_{i \in \{1,..,4\}}$, which can be easily tuned to minimize the regret using, for instance, Bayesian optimization. For hyper-parameter tuning of oGGT, we have utilized Bayesian optimization instead of Line-search, while keeping all other parts of Algorithm~\ref{HPtun} the same.
	
	In summary, the proposed oGGT algorithm is optimized for a given network through the specified $\W$ and for the given loss function, through $L$ and $\M$. The process is summarized in Fig. \ref{fig:diag_hyp_prt_tng}.
	
	\setlength{\unitlength}{1cm}
	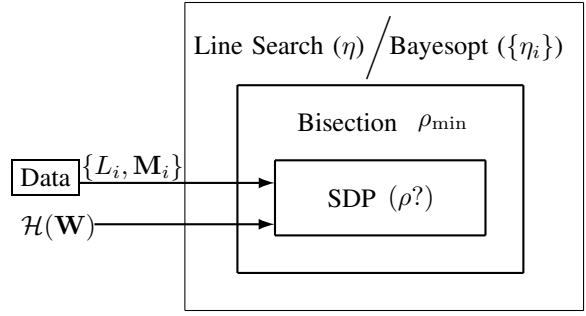
\begin{figure}
		\centering
		\begin{picture}(7,4.2) 
			
			\put(2.3,0){\line(0,1){4.1}}  
			\put(2.3,4.1){\line(1,0){5.3}}
			\put(7.6,4.1){\line(0,-1){4.1}}
			\put(7.6,0){\line(-1,0){5.3}}
			
			\thicklines
			\put(3,0.5){\line(0,1){2.5}} 
			\put(3,3){\line(1,0){3.8}}
			\put(6.8,3){\line(0,-1){2.5}}
			\put(6.8,0.5){\line(-1,0){3.8}}
			
			\put(3.5,1){\line(0,1){1}} 
			\put(3.5,2){\line(1,0){2.8}}
			\put(6.3,2){\line(0,-1){1}}
			\put(6.3,1){\line(-1,0){2.8}}
			
			\put(0.90,1.7){\vector(1,0){2.6}}
			\put(0.8,1.9){\makebox(0,0)[l]{ $\{L_i,\M_i\}$}}
			
			\put(1.1,1.15){\vector(1,0){2.4}}
			\put(0.005,1.1){\makebox(0,0)[l]{ $\cH (\W)$}}
			
			\put(0.0001,1.65){\framebox{Data}}
			\put(2.3,3.5){\makebox(0,0)[l]{\ Line Search ($\eta$)\bigg/}}
			\put(4.9,3.5){\makebox(0,0)[l]{\ Bayesopt ($\{\eta_i\}$)}}
			\put(3.8,2.5){\makebox(0,0)[l]{Bisection}}
			\put(5.4,2.5){\makebox(0,0)[l]{$\rho_{\min}$}}
			\put(4.2,1.5){\makebox(0,0)[l]{SDP}}
			\put(5.0,1.5){\makebox(0,0)[l]{$(\rho ?)$}}
		\end{picture}
		\caption{Block diagram for hyperparameter tuning of GGT}.
		\label{fig:diag_hyp_prt_tng}
	\end{figure}

	\section{Numerical Analysis}
	\label{secnumanalysis}
	In this section, we demonstrate the superior performance of the proposed GGT algorithms, and in particular that of the oGGT algorithm, as compared to the state-of-the-art decentralized online algorithms, namely DOG \cite{wu2022decentralized}, DDAG \cite{nazari2021adaptive}, DOO-GT \cite{zhang2019distributed}, D-OCO \cite{zhang2020distributed}, and GTAdam \cite{carnevale2022gtadam}. We consider a target tracking application using synthetic data and an online learning problem using the room-occupancy data set from \cite{candanedo2016accurate}. It is remarked that we do not include the performance of DMD \cite{shahrampour2018distributed} but only its Euclidean version DOG \cite{wu2022decentralized}. Further, the performance of DMD is known to be worse than that of  D-OCO \cite{zhang2020distributed}, which is a special case of our algorithm. 
	
	\subsection{Tracking Time Varying Target via Least-Squares} \label{TT}
	We consider a fully connected network with $n = 10$ and generate its mixing matrix $\W$ using the distributed Metropolis–Hastings (MH) algorithm with transition probability parameter $p=0.6$, as described in \cite[Appendix G]{dixit2020online}. The nodes in the network represent sensors that collaborate to track three two-dimensional sinusoidal signals given by
	\begin{align*}
		\w_j^k &=
		\begin{bmatrix}
			A_j \sin(\omega k + \phi_j) \\
			\omega A_j \cos(\omega k + \phi_j)
		\end{bmatrix}
	\end{align*}
	for $j \in \{1,2,3\}$ and $k \geq 1$, representing positions and velocities of three moving targets. The signal parameters $\{A_j, \phi_j\}$ are uniformly generated from the intervals $[0.4,1]$ and $[0,\pi]$, respectively, while we use $\omega = 4\pi$ radians. 
	At each $k$, the three signal vectors are collected into the super-vector $\s^k$. The $i$-th sensor node receives measurements $\v_i^k = \C_i\w^k \in \mathbb{R}^4$, where $\C_i \in \mathbb{R}^{6 \times 6}$ contains randomly distributed entries from the range $[0,1]$. While generating these observation matrices, we reject those that do not satisfy $\lambda_{\min}(\C_i^\T\C_i) < 1.5$. In order to track $\s^k$, each node considers the least-squares objective $f_i^k(\x) = \frac{1}{2}\norm{\C_i\x - \v_i^k}^2$ as its local loss function. As the data is distributed among nodes, all the nodes must coordinate to minimize the global objective in \eqref{fav}.   
	
\begin{figure}
		\centering
		\includegraphics[scale=1]{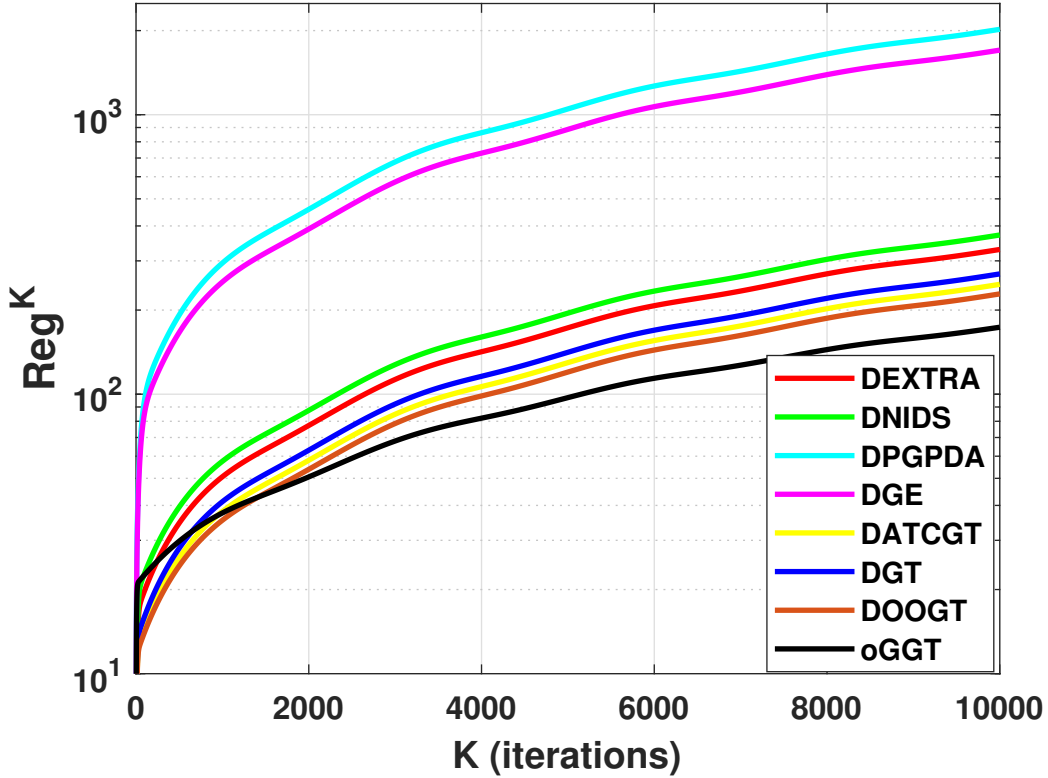}
		\caption{Regret rates of the GGT variants for target tracking}
		\label{fig:regret}
	\end{figure}

	The objective function parameters $\mu_i=\lambda_{\min}(\C_i^{\mathsf{T}}\C_i)$ and $L_i=\lambda_{\max}(\C_i^{\mathsf{T}}\C_i)$ are first used to tune the hyperparameters of all the GGT variants as well as oGGT, as detailed in Sec. \ref{tuning}. 
	The hyperparameters of the other state-of-the-art algorithms (DOG, DDAG, GTAdam) are tuned experimentally to yield the best possible performance. 
	\begin{figure}
		\centering
		\includegraphics[scale=1]{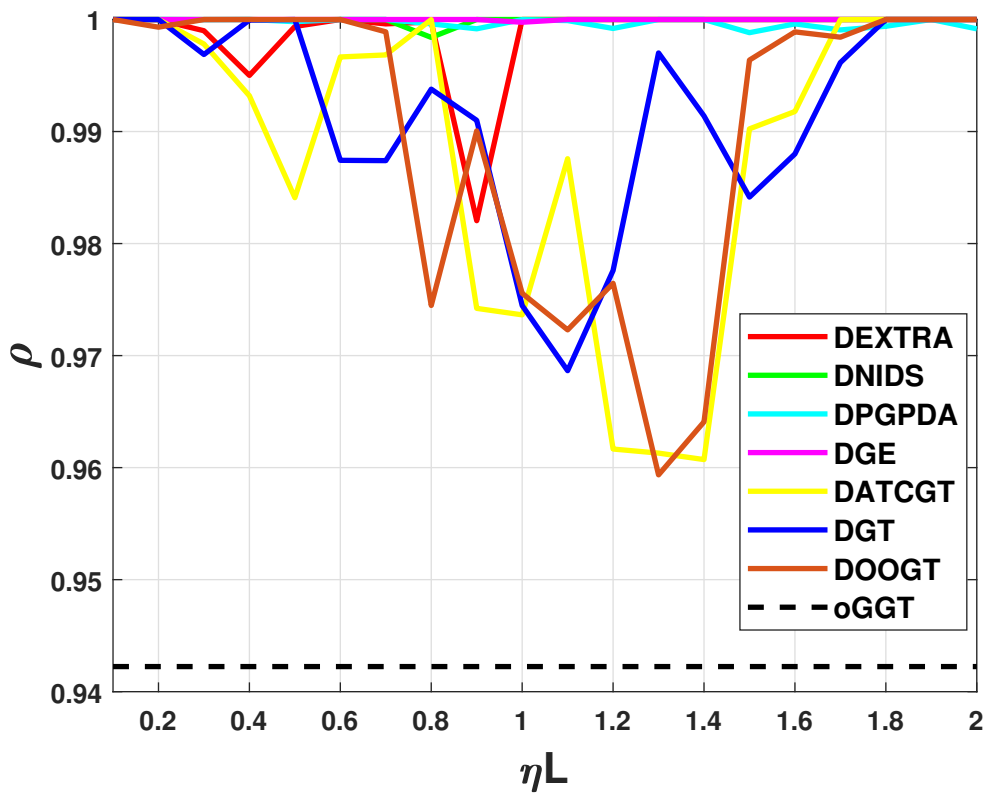}
		\caption{Convergence rate parameter ($\rho$) of various algorithms against the step-size parameter ($\eta$). The dashed line for oGGT corresponds to the optimal $\rho$ obtained after tuning all its hyperparameters.}
		\label{fig:prmtr_tng}
	\end{figure} 
	Fig. \ref{fig:prmtr_tng} shows the variation of the best possible feasible values of $\rho$ against the step-sizes. It can be seen that most of the variants are sensitive to the choice of the step-size. The figure also shows the final value of $\rho$ obtained for oGGT, which is better than best possible value of $\rho$ obtained for each variants, suggesting that the performance of oGGT is likely to be superior to that of the other variants. This is verified in Fig. \ref{fig:regret}, which shows the empirical value of the regret rate obtained for the different variants as well as that of oGGT. The figure also includes the performances of DOO-GT \cite{zhang2019distributed} and D-OCO \cite{zhang2020distributed}, which is equivalent to D-GT. 
	
	\begin{figure}
		\centering
		\includegraphics[scale=1]{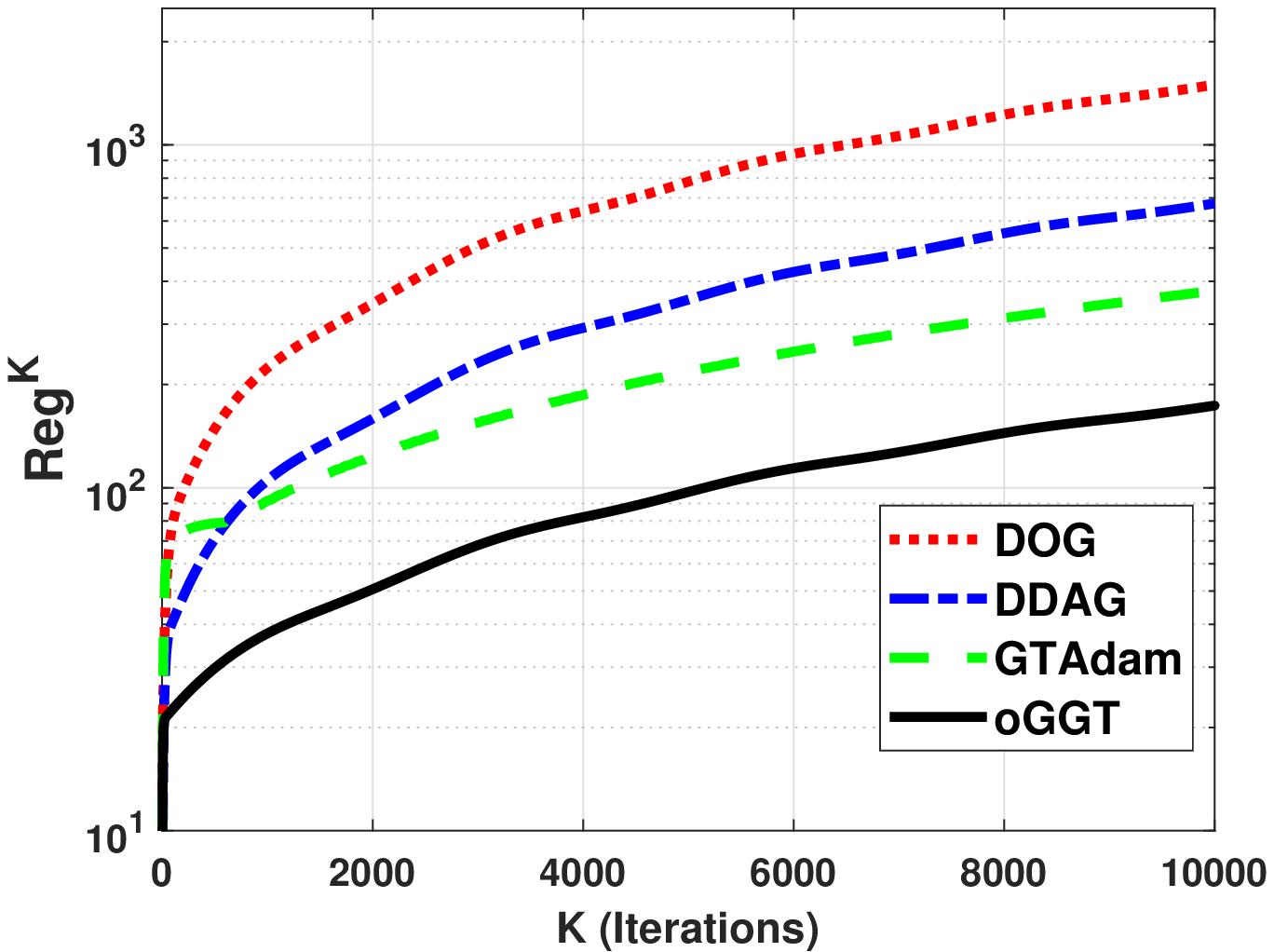}
		\caption{Regret rates of state-of-the-art algorithms and oGGT for target tracking}
		\label{fig:reg_SOTA}
	\end{figure}
	Fig. \ref{fig:reg_SOTA} shows the superior performance of oGGT as compared to the state-of-the-art algorithms DOG, DDAG, and GTAdam. We do not include the performance of  DOO-GT and D-OCO, both of which are special cases of GGT and already compared in Fig. \ref{fig:regret}. Here, DOG and DDAG do not employ gradient tracking, and hence have highest regret rates. While GTAdam does employ momentum, it still falls short of the proposed oGGT.  
	
	\subsection{Real Data : Online learning } \label{sec:num_ana_onl9_ler}
	We consider the problem of predicting room occupancy using data from sensors measuring temperature, relative humidity, ambient light, carbon dioxide concentration, and humidity ratio. The experiment utilizes the Room Occupancy Detection dataset from \cite{candanedo2016accurate} containing 20560 data points of the form $\{\a_i,y_i\}$ where $\a_i\in \Rn^5$ are the standardized sensor measurements and $y_i \in\{-1,1\}$ are labels. For the purpose of tracking room occupancy, we consider a network of five nodes, each of which receive a subset of data points, and seek to minimize the regularized logistic loss $f_i^k(\x)=\log(1+\exp(-y_i^k(\a_i^k)^\T\x))+\frac{\gamma}{2}\norm{\x}^2$ with $\gamma = 10^{-2}$. For the purpose of decentralized optimization, we consider a random network and generated $\W$ using the MH algorithm with $p = 0.8$. 
	
	We assume that $10\%$ of the data is available offline and can be used for tuning the various hyperparameters. In particular, we use the offline data to estimate $\{L_i,\M_i\}$ for the GGT variants \cite[Preposition 3.1]{freund2018condition} and the hyperparameters of the state-of-the-art algorithms. 
	\begin{figure}
		\centering
		\includegraphics[scale=0.8]{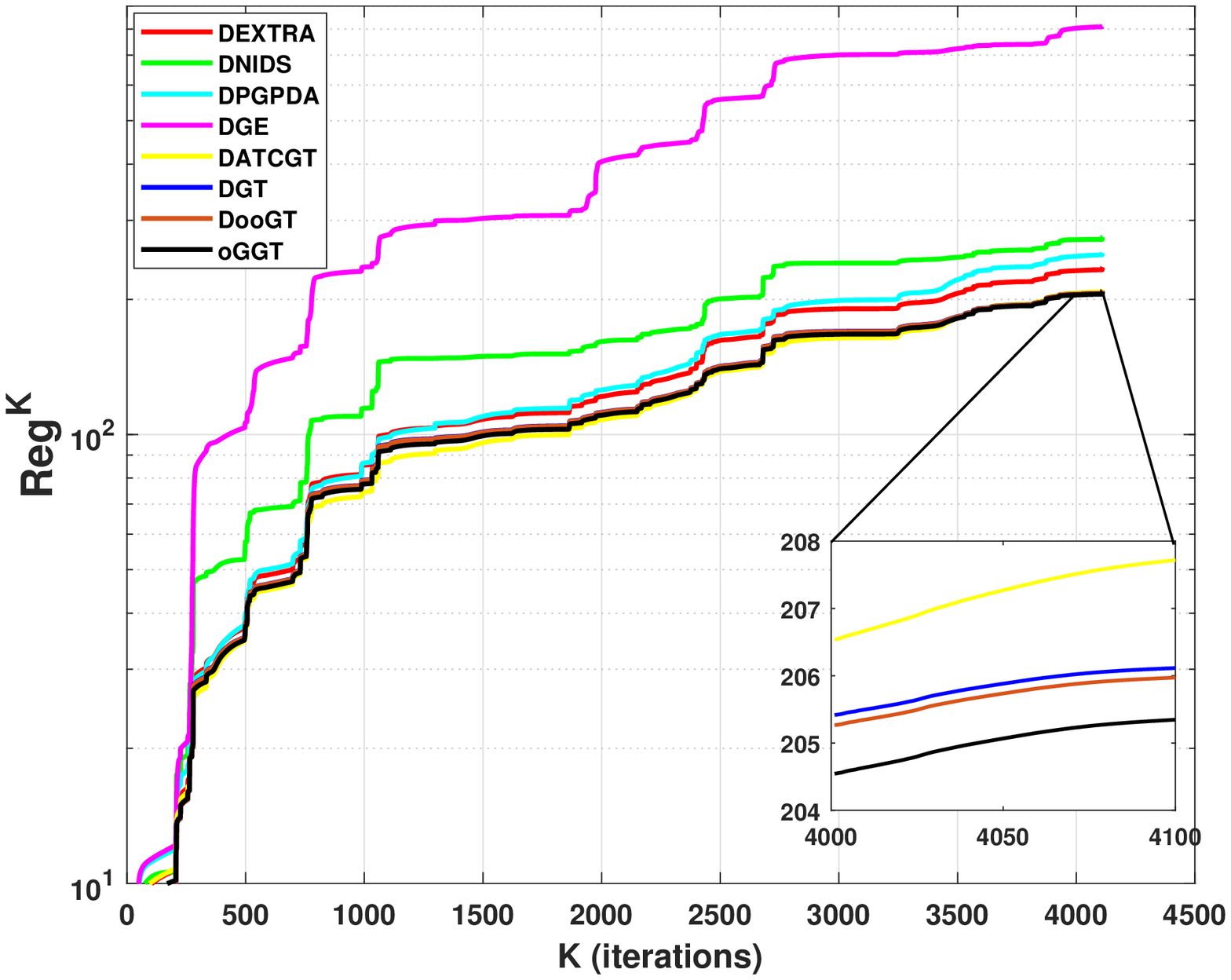}
		\caption{ Regret rates of the GGT variants for room occupancy prediction}
		\label{fig:Room_occ}
	\end{figure}
	\begin{figure}
		\centering
		\includegraphics[scale=1]{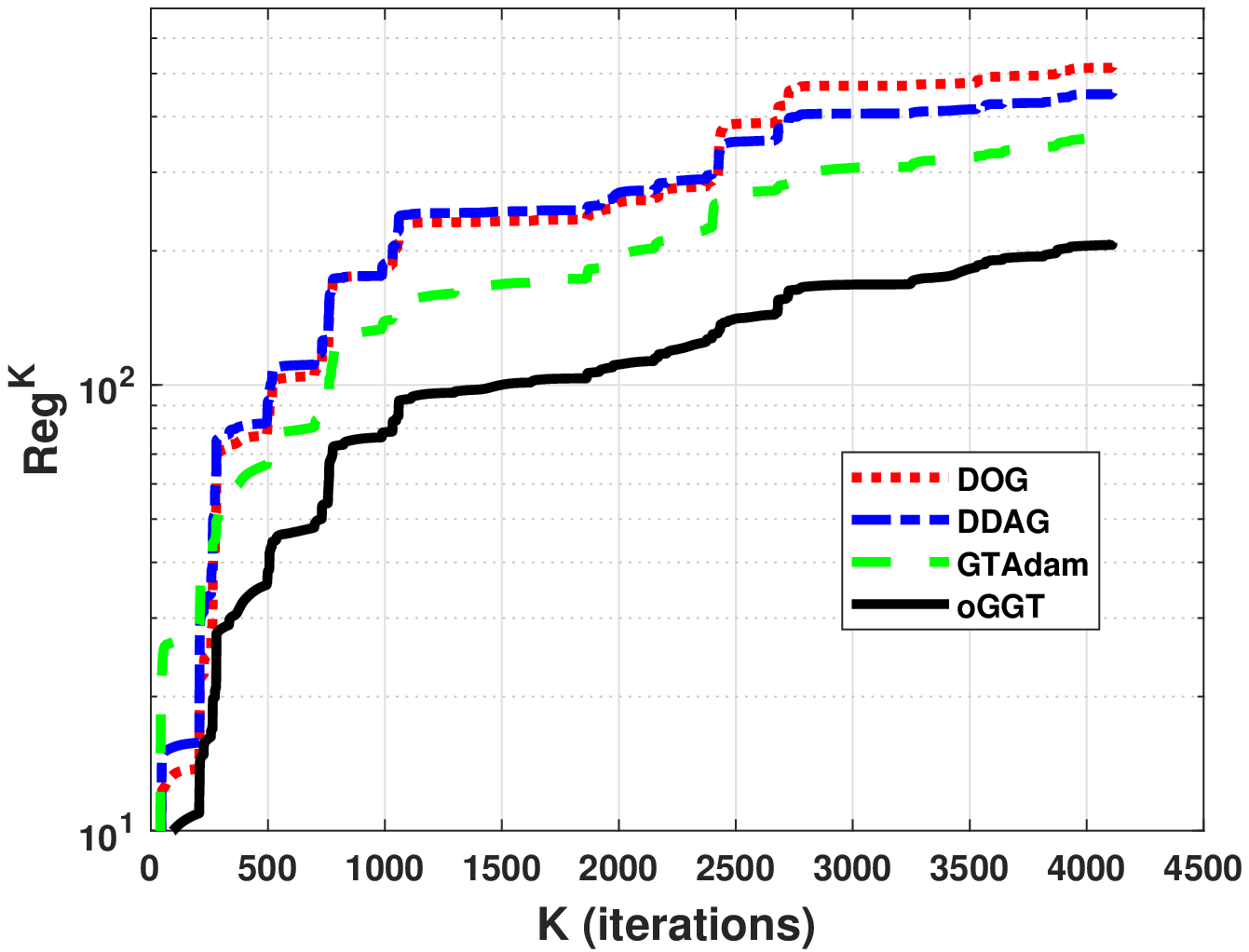}
		\caption{ Regret rates of the state-of-art algorithms and oGGT for room occupancy prediction}
		\label{fig:RO_regret_SOTA}
	\end{figure}	
	
	Figure \ref{fig:Room_occ} shows the performance of various GGT variants as well as that of oGGT, which as expected, performs the best. Fig. \ref{fig:RO_regret_SOTA} also establishes the significantly better performance of oGGT as compared to the other state-of-the-art algorithms.

	\section{Conclusion}
	\label{secconclusion}
	This work considered the decentralized online learning problem that involves  minimization of sum of time-varying functions spread across multiple disparate nodes. To solve the problem, we propose the Generalized Gradient Tracking (GGT) framework that unifies several existing approaches. The proposed algorithm is analyzed using a novel semidefinite programming-based approach that allows us to drop the assumption on gradient boundedness, commonly found in many existing works. The regret bounds improve upon similar bounds that have been separately developed for the different special cases of GGT, and provide fresh bounds for several novel variants of GGT based on classical decentralized algorithms. Additionally, we put forth an optimized version of GGT whose parameters can be tuned offline using only the problem parameters, resulting in superior dynamic regret performance as well as superior empirical performance. The proposed framework can be viewed as a template for designing optimal algorithms in general, and showcases the importance of designing algorithms that are both theoretically and empirically optimal. 
	
	\appendices
	\section{Proof of equation~\eqref{initialz_cmpct}}\label{pr:eq:initialz_cmpct}
	Using \eqref{compact} on the left-hand side of \eqref{initialz_cmpct} we get 
	\begin{align}
		&\H^{(6)}\ux^{k-1} + \H^{(7)} \us^{k-1} + \H^{(8)}\nab^{k-1} \nonumber \\
		&= \H^{(6)} \big( \H^{(1)}\ux^{k-2} + \H^{(2)}\us^{k-2} \big) + \H^{(8)}\nab^{k-1} + \H^{(7)} \big(\H^{(3)}\ux^{k-2} + \H^{(4)} \us^{k-2} + \H^{(5)} (\nab^{k-2} - \nab^{k-1}) \big),  \nonumber \\
		&= (\H^{(6)} \H^{(1)} +\H^{(7)} \H^{(3)}) \ux^{k-2} + \H^{(7)} \H^{(5)}\nab^{k-2} + (\H^{(6)} \H^{(2)} +\H^{(7)} \H^{(4)} ) \us^{k-2} + (-\H^{(7)}\H^{(5)} + \H^{(8)}) \nab^{k-1}. \nonumber 	
	\end{align}
	Further, using \eqref{Mequalities} we get
	\begin{align}
		\H^{(6)}&\ux^{k-1} + \H^{(7)} \us^{k-1} + \H^{(8)}\nab^{k-1} = \H^{(6)} \ux^{k-2} + \H^{(7)}\us^{k-2} + \H^{(8)} \nab^{k-2}. \nonumber 	
	\end{align}
	Recursively applying, \eqref{compact} for $k-2,k-3,\ldots,1$ and \eqref{Mequalities}, we obtain:
	\begin{align}
		\H^{(6)}&\ux^{k-1} + \H^{(7)} \us^{k-1} + \H^{(8)}\nab^{k-1} = \H^{(6)}\ux^{0} + \H^{(7)} \us^{0} + \H^{(8)}\nab^{0}, \overset{\eqref{initialz_cmpct0}}{=}  \O_{d \times 1}, \nonumber 
	\end{align}
	which yields the desired result.
	
	\section{Updates for the proposed dynamic algorithms}\label{updates}
	The definitions of $\z^k$ and $\z^{k,\star}$ are identical for D-EXTRA, D-NIDS, and D-PGPDA. Similar to D-EXTRA, we can analyze D-NIDS and D-PGPDA. For all the algorithms, the initial $\x_i^0$ is arbitrary. The D-NIDS updates take the form:
	\begin{align}
		\label{D-NIDSupdate}
		\x_i^{k+1}= &\x_i^{k}+\sum_{j=1}^{n} W_{ij} \x_j^{k} - 0.5\x_i^{k-1} - 0.5\sum_{j=1}^{n} 	W_{ij} \x_j^{k-1} -\frac{\eta_i}{2} \nabla f_{i}^{k}(\x_i^{k}) + \frac{\eta_i}{2} \nabla 	f_{i}^{k-1}(\x_i^{k-1}) \nonumber\\
		&-\frac{1}{2}\sum_{j=1}^{n} \eta_j \bigg{(} W_{ij}\nabla f_j^{k}(\x_j^{k}) -W_{ij}\nabla f_j^{k-1}(\x_j^{k-1})\bigg{)},
	\end{align}
	where $\x_i^1=\x_i^0-\eta_i \nabla f_i^0(\x_i^0)$ for all $1\leq i \leq n$.  
	The D-PGPDA updates take the form
	\begin{align}
		\label{D-PGPDA}
		\x_i^{k+1}&=\x_i^k- \frac{1}{2 \eta d_i} (\nabla f_i^k(\x_i^k)-\nabla 	f_i^{k-1}(\x_i^{k-1})) + \sum_{j \in \cN(i)}\frac{1}{d_i}\x_j^k-\frac{1}{2} \bigg( \sum_{j \in \cN(i)}\frac{1}{d_i}\x_j^{k-1}+x_i^{k-1}\bigg),
	\end{align} 
	where $d_i$ represents the degree of node $i$ and $\x_i^1=\x_i^0-\frac{1}{2 \eta d_i} \nabla f_i^0(\x_i^0)$ for all $1\leq i \leq n$.
	Next, we will establish a connection between GGT and the D-GE algorithm. By defining $\s_{i}^{k-1} = \g_{i}^{k-1}$ and initializing $\g_i^0=-\nabla f_i^0(\x_i^0)$ for all $i \in \cV$, the updates described in \eqref{gt1}-\eqref{gt2} can be expressed as follows:
	\begin{align}
		\label{DQuLi}
		\x_i^{k} &= \sum_{(i,j) \in \cE} W_{i,j}\x_j^{k-1} + \eta \g_i^{k-1}, \nonumber \\
		\g_i^{k} &= \sum_{(i,j) \in \cE} W_{i,j}\g_j^{k-1}  - \nabla f_{i}^{k}(\x_{i}^{k}) + 	\nabla f_{i}^{k-1}(\x_{i}^{k-1}). 
	\end{align}
	Here, $W_{i,j}$ represents the elements of a doubly stochastic matrix, and $W_{i,j} > 0$ if and only if $(i,j) \in \cE'$. The update equations for D-GE closely resemble those in \eqref{xi-up}-\eqref{si-up}, and therefore, D-GE can be expressed as a specific case of GGT by defining $\cH$ as shown in Table~\ref{newformtable}. Following a similar approach, we can summarize the updates for D-GT, D-ATC-GT, DOOGT, and oGGT. The updates for D-ATC-GT take the form: 
	\begin{align}
		\label{D-ATC-GT}
		\x_i^{k}&=\sum_{j=1}^{n} W_{i,j} \left(\x_j^{k-1} + \eta \g_i^{k-1}\right), \nonumber \\
		\g_i^{k}&=\sum_{j=1}^{n} W_{i,j} \left(\g_j^{k-1}-\nabla f_{i}^{k}(\x_{i}^{k})+\nabla 	f_{i}^{k-1}(\x_{i}^{k-1})\right),
	\end{align}
	where $\g_i^0=-\nabla f_i^0(\x_i^0)$ for all $1\leq i \leq n$. The updates for D-GT take the form:
	\begin{align}
		\label{D-GT}
		\x_i^{k}&=\sum_{j=1}^{n} W_{i,j}\bigg(\x_j^{k-1}+\g_j^{k-1} \bigg), \nonumber \\
		\g_i^{k}&=\sum_{j=1}^{n} W_{i,j}\g_j^{k-1}- \eta \nabla f_{i}^{k}(\x_{i}^{k})+\eta \nabla f_{i}^{k-1}(\x_{i}^{k-1}),
	\end{align}
	where $\g_i^0=-\eta \nabla f_i^0(\x_i^0)$ for all $1\leq i \leq n$. 
	Finally, the updates for oGGT take the form:
	\begin{align}
		\label{oGGT}
		\x_i^{k}&=\sum_{j=1}^{n} W_{i,j} \bigg(\x_j^{k-1}+ \beta \g_j^{k-1} \bigg) + \alpha 	\g_i^{k-1}, \nonumber \\
		\g_i^{k}&=\sum_{j=1}^{n} W_{i,j} \bigg( \g_j^{k-1} + \delta \big( \nabla 	f_{j}^{k-1}(\x_{j}^{k-1}) - \nabla f_{j}^{k}(\x_{j}^{k}) \big) \bigg) + \gamma \big( \nabla f_{i}^{k-1}(\x_{i}^{k-1}) - \nabla f_{i}^{k}(\x_{i}^{k}) \big), 
	\end{align}
	where $\g_i^0=-(\gamma + \delta) \nabla f_i^0(\x_i^0)$ for all $1\leq i \leq n$.

	\section{Proof of Corollary~\ref{col: zk1_zs_bnd}}\label{col_pr: zk1_zs_bnd}
	\begin{IEEEproof}
		We begin with bounding $\Vert \z^{k+1}-\z^{k,\star}\Vert_\P$ via repeated application of the result of Theorem~\ref{mytheo1}, which says that
		\begin{align}
			\Vert \z^{k+1}&-\z^{k,\star}\Vert_\P \leq \rho \Vert \z^{k}-\z^{k,\star}\Vert_\P, \\
			&=\rho \Vert \z^k-\z^{k-1,\star}+\z^{k-1,\star}-\z^{k,\star}\Vert_\P, \\
			&\leq  \rho \Vert \z^k-\z^{k-1,\star}\Vert_\P + \rho \Vert 	\z^{k,\star}-\z^{k-1,\star}\Vert_\P, \label{triangle1}\\
			&\lfrom{zbound} \rho \Vert \z^k -\z^{k-1,\star} \Vert_\P + \rho \sigma, \label{rec1}
		\end{align} 
		where we have introduced $\z^{k-1,\star}$ and applied the triangle inequality in \eqref{triangle1}. Recursively applying \eqref{rec1} for $k$, $k-1$, $\ldots$, 1, we obtain:
		\begin{align}
			\Vert \z^{k+1}&-\z^{k,\star}\Vert_\P \leq \rho \Vert \z^k -\z^{k-1,\star} \Vert_\P + \rho \sigma, \\
			&\leq \rho^2 \Vert \z^{k-1} -\z^{k-2,\star} \Vert_\P + \rho(1+\rho) \sigma, \\
			&\leq \rho^{k-1}\Vert \z^{2} -\z^{1,\star} \Vert_\P + \rho(1+\ldots+\rho^{k-2}) \sigma, \\
			&\leqtext{Th. \ref{mytheo1}} \rho^k\Vert \z^{1} -\z^{1,\star} \Vert_\P + \frac{\sigma\rho(1-\rho^{k-1})}{1-\rho}.\label{cor1ineq}
		\end{align}
		Finally, we can write \eqref{cor1ineq} in terms of $\ell_2$ norm by using the property that $\sqrt{\lambda_{\min}(\P)}\norm{\x} \leq \norm{\x}_{\P} \leq \sqrt{\lambda_{\max}(\P)}\norm{\x}$ for any vector $\x$, which yields the required result. 
	\end{IEEEproof}
	
	\section{Proof of Corollary \ref{col:grad_bnd}}\label{col_pr:grad_bnd}
	\begin{IEEEproof}
		We begin with bounding $\norm{\x_i^k - \xks}$ in terms of $\norm{\z^{k+1}-\z^{k,\star}}$. As per the definitions of $\z^k$ and $\z^{k,\star}$ in Table~\ref{defznzs}. $\z^k$ contains $\x_i^{k-1}$ and the corresponding component of $\z^{k,\star}$ is $\xks$ (see Table~\ref{defznzs}) hence, for each user $i$, $\Vert \x_i^k-\xks \Vert \leq \Vert \z^{k+1}-\z^{k,\star} \Vert$. Further utilizing Theorem~\ref{mytheo1} and Corollary~\ref{col: zk1_zs_bnd}, we can bound $\Vert \x_i^k-\xks \Vert$ as below 
		\begin{align}
			\Vert \x_i^k-\xks \Vert &\leq \Vert \z^{k+1}-\z^{k,\star}, \Vert \nonumber \\
			&\leq\frac{1}{\sqrt{\lambda_{\min}(\P)}}\Vert \z^{k+1}-\z^{k,\star} \Vert_\P, 	\tag{Property of $\P$-norm}\nonumber \\
			&\leq \frac{\rho}{\sqrt{\lambda_{\min}(\P)}}\Vert \z^k-\z^{k,\star} \Vert_\P, 	\tag{Theorem~\ref{mytheo1}} \nonumber \\ 
			&\overset{(a)}{\leq} \frac{\rho^k}{\sqrt{\lambda_{\min}(\P)}} \Vert \z^1-\z^{1,\star}  	\Vert _\P+\frac{\sigma \rho (1-\rho^{k-1})}{\sqrt{\lambda_{\min}(\P)}(1-\rho)}.\label{xik_xks_bound}
		\end{align} 
		Where in (a) we have utilized Corollary~\ref{col: zk1_zs_bnd}.
		As, $f^k$ is smooth (Lipschitz gradient) with parameter $L$ for all $k=1,2,3,...$, we have
		\begin{equation*}
			\Vert \nabla f^k(\x_i^k) -  \nabla f^k(\xks)\Vert \leq L\Vert \x_i^k-\xks \Vert.
		\end{equation*} 
		Also, we have $\nabla f^k(\xks)=0$ and $\Vert \x_i^k-\xks \Vert$ is bounded for each time instance $k$. Finally we can bound $\norm{\nabla f^k(\x_i^k)}$ using \eqref{xik_xks_bound}, which yields the required result. 
	\end{IEEEproof}

	\section{Proof of Lemma \ref{lem:wk_ck_dk}} \label{lem_pr:wk_ck_dk}
	\begin{IEEEproof}
		From the definition of $W_K$ and using the property of $\P$-norm, we have that
		\begin{align}
			W_K &= \sum_{k=2}^{K} \Vert \z^{k,\star}-\z^{k-1,\star} \Vert_{\P}, \nonumber \\
			&\leq \sqrt{\lambda_{\max}(\P)}\sum_{k=2}^{K} \Vert \z^{k,\star}-\z^{k-1,\star} \Vert. \label{eq_pro_pr:W_K}
		\end{align} 
		As per definition \eqref{zdef}, the state vector $\z^k$ includes the optimization vector $\ux^{k-1}$, gradient vector $\nab^k$ and $\us^{k-1}$. Also, $\us^{k-1}$ is the combination of $\ux^{k-1}$ and $\nab^k$, for example in D-EXTRA $\s^{k-1}=\x^{k}$, hence the optimal state vector $\z^{k,\star}$ can only have $\ux^{k,\star} \in \Rn^{nd \times 1} $ , $\nab^{k,\star} \in \Rn^{nd \times 1}$ as its component vectors. We can split  $\z^k$ as follow  
		\begin{align}
			\Vert \z^{k,\star} &-\z^{k-1,\star} \Vert^2 \nonumber \\
			&\leq 2 \Vert \ux^{k,\star}-\ux^{k-1,\star} \Vert^2 +2\Vert 	\nab^{k,\star}-\nab^{k-1,\star} \Vert^2, \label{zkreltdckdk} \\
			&\leq 2 \left(\Vert \ux^{k,\star}-\ux^{k-1,\star} \Vert + \Vert \nab^{k,\star}-\nab^{k-1,\star} \Vert \right)^2.\nonumber
		\end{align}
		On taking square root of the above equation and substituting in \eqref{eq_pro_pr:W_K}, we obtain
		\begin{align}
			W_K &\leq \sqrt{\lambda_{\max}(\P)}  \sum_{k=2}^{K}  \Vert \ux^{k,\star}-\ux^{k-1,\star} 	\Vert + \sqrt{\lambda_{\max}(\P)}  \sum_{k=2}^{K} \Vert \nab^{k,\star}-\nab^{k-1,\star} \Vert \nonumber.
		\end{align}
		From the property of norm and the definition of $C_K$ \eqref{ck} and $D_K$ \eqref{dk} we obtain $W_K \leq \sqrt{2n\lambda_{\max}(\P)} ( C_K + D_K)$. For second relation, from the definition of $W_{K,2}$ and using property of $\P$-norm we have
		\begin{align}
			W_{K,2} &= \sum_{k=2}^{K} \Vert \z^{k,\star}-\z^{k-1,\star} \Vert_{\P}^2, \nonumber \\
			&\leq \lambda_{\max}(\P) \sum_{k=2}^{K} \Vert \z^{k,\star}-\z^{k-1,\star} \Vert^2. \label{eq_pro_pr:W_K2}
		\end{align} 
		Utilizing the bound obtained in \eqref{zkreltdckdk} on $\Vert \z^{k,\star}-\z^{k-1,\star} \Vert^2$ we obtain
		\begin{align}
			W_{K,2} &\leq \lambda_{\max}(\P) \sum_{k=2}^{K} 2 \Vert \ux^{k,\star} - \ux^{k-1,\star} \Vert^2 + \lambda_{\max}(\P) \sum_{k=2}^{K} 2\Vert \nab^{k,\star} - \nab^{k-1,\star} \Vert^2 .
		\end{align}
		From the property of norm and the definition of $C_{K,2}$ and $D_{K,2}$, we obtain the desired result.
	\end{IEEEproof}
	
	\section{Proof of Lemma \ref{lem2}}\label{pr_lem2}
	\begin{IEEEproof}
		We begin with expanding the summation on the left hand side of \eqref{linear bound} as below	
		\begin{align*}
			\sum _{k=1}^{K} \Vert \z^k-\z^{k, \star} \Vert_\P = \Vert \z^1 - \z^{1, \star} \Vert_\P + \sum _{k=2}^{K} \Vert \z^k-\z^{k, \star} \Vert_\P. 
		\end{align*}
		Now on introducing $\z^{k-1,\star}$ in the second term, applying triangle inequality and Theorem~\ref{mytheo1} we obtain
		\begin{align}
			\sum _{k=1}^{K} \Vert \z^k-\z^{k, \star} \Vert_\P &\leq  \Vert \z^1 - \z^{1, \star} 	\Vert_\P +   \sum _{k=2}^{K} \Vert \z^k-\z^{k-1, \star} \Vert_\P + \sum _{k=2}^{K} \Vert \z^{k,\star}-\z^{k-1, \star} \Vert_\P,\\
			&\leq  \Vert \z^1 - \z^{1, \star} \Vert_\P +   \sum _{k=2}^{K} \rho \Vert 	\z^{k-1}-\z^{k-1, \star} \Vert_\P +  \sum _{k=2}^{K} \Vert \z^{k,\star}-\z^{k-1, \star} \Vert_\P. \label{lem2_pr_refeq}
		\end{align}
		On adding and subtracting $\rho \Vert \z^{K}-\z^{K, \star} \Vert_\P$ and updating the limits of summation in right hand side of \eqref{lem2_pr_refeq}, we get
		\begin{align}
			\sum _{k=1}^{K} \Vert \z^k-&\z^{k, \star} \Vert_\P   
			\leq   \Vert \z^1 - \z^{1, \star} \Vert_\P +   \sum _{k=1}^{K} \rho \Vert 	\z^{k}-\z^{k, \star} \Vert_\P -\rho \Vert \z^{K}-\z^{K, \star} \Vert_\P + \sum _{k=2}^{K} \Vert \z^{k,\star}-\z^{k-1, 	\star} \Vert_\P. \label{lem2_pr_refeq2} 
		\end{align}
		On subtracting $\rho \sum _{k=1}^{K} \Vert \z^{k}-\z^{k, \star} \Vert_\P $ from both side of \eqref{lem2_pr_refeq2}and doing some mathematical manipulations, utilizing the fact that $\rho \in (0,1)$ we get
		\begin{align}
			\sum _{k=1}^{K} \Vert \z^k-\z^{k, \star} \Vert_\P  
			&\leq  \frac{1}{1-\rho} \Vert \z^1 - \z^{1, \star} \Vert_\P - \frac{\rho}{1-\rho} \Vert 	\z^{K}-\z^{K, \star} \Vert_\P + \frac{1}{1-\rho} \sum _{k=2}^{K} \Vert \z^{k,\star}-\z^{k-1, \star} \Vert_\P. \label{lem2_pr_refeq3} 
		\end{align}
		Now, utilizing the definition of $W_K$ (\eqref{pathlength}, p=1) and discarding the second term of R.H.S. of \eqref{lem2_pr_refeq3} as $\frac{\rho}{1-\rho} \Vert \z^{K}-\z^{K, \star} \Vert_\P $ is always positive, we get the desired result \eqref{linear bound}. 
		
		Similarly for the second relation, we begin with expanding the summation on the left hand side of \eqref{quad_bound} as below	
		\begin{align*}
			\sum_{k=1}^K  \norm{\z^{k} - \z^{k,\star} }_{\P}^2 &= \norm{\z^{1} - \z^{1,\star} 	}_{\P}^2 + \sum_{k=2}^K  \norm{\z^{k} - \z^{k,\star} }_{\P}^2.
		\end{align*}
		Introducing $\z^{k-1,\star}$ in the second term, applying Young's inequality with $\alpha_1 = \frac{1 -\rho^2}{2 \rho^2}$ we obtain
		\begin{align}
			&\sum_{k=1}^K  \norm{\z^{k} - \z^{k,\star} }_{\P}^2 \leq \norm{\z^{1} - \z^{1,\star} 	}_{\P}^2 + \sum_{k=2}^K \Bigg( \frac{1 + \rho^2}{2 \rho^2} \norm{\z^{k} - \z^{k-1,\star}}_{\P}^2 	+ \frac{1+\rho^2}{1-\rho^2} \norm{\z^{k-1,\star}- \z^{k,\star} }_{\P}^2 \Bigg). \nonumber
		\end{align}
		From Theorem~\ref{mytheo1} we have
		\begin{align}
			\sum_{k=1}^K & \norm{\z^{k} - \z^{k,\star} }_{\P}^2 
			\leq \norm{\z^{1} - \z^{1,\star} }_{\P}^2 +  \frac{1 + \rho^2}{2 } 	\sum_{k=1}^K  \norm{\z^{k} - \z^{k,\star}}_{\P}^2 + 	\frac{1+\rho^2}{1-\rho^2} \sum_{k=2}^K  \norm{\z^{k-1,\star}- \z^{k,\star} }_{\P}^2. \nonumber
		\end{align}
		Taking the second term on the right to the left, we obtain
		\begin{align}
			\frac{1 - \rho^2}{2 }\sum_{k=1}^K  &\norm{\z^{k} - \z^{k,\star} }_{\P}^2 \leq 		\norm{\z^{1} - \z^{1,\star} }_{\P}^2 + \frac{1+\rho^2}{1-\rho^2} \sum_{k=2}^K  \norm{\z^{k-1,\star}- \z^{k,\star} }_{\P}^2. \nonumber
		\end{align}
		Finally, since $\rho \in (0,1)$, we obtain the bound
		\begin{align}
			\sum_{k=1}^K  \norm{\z^{k} - \z^{k,\star} }_{\P}^2 	&\leq \frac{2}{1 - \rho^2} 		\norm{\z^{1} - \z^{1,\star} }_{\P}^2 + \frac{4}{(1-\rho^2)^2} W_{K,2}. \nonumber
		\end{align}
	\end{IEEEproof}
	
	\section{Dependence of Regret on $\P$ and $\rho$}\label{pr_cond_P_rho}
	Here, we bound the regret so that its dependence on $\P$ and $\rho$ is explicit. Utilizing \eqref{eq:lem:wk2_ck2_dk2} and Theorem~\ref{mytheo3} we have
	\begin{align}
		\Reg^K &\leq \frac{2L \rho^2}{n \lambda_{\min}(\P) (1-\rho^2)^2} \Bigg( \norm{\z^{1} - 	\z^{1,\star} }_{\P}^2 + 2 \lambda_{\max}(\P) ( C_{K,2} +  D_{K,2}) \Bigg), \nonumber \\
		&\leq \frac{4L \rho^2 \text{cond}(\P)}{ (1-\rho^2)^2} \Bigg( \norm{\z^{1} - \z^{1,\star} 	}^2 + C_{K,2} +  D_{K,2} \Bigg), \nonumber \\
		&\overset{(a)}{\leq} \frac{4 L \rho^2 \text{cond}(\P)}{ (1-\rho)^2} ( \norm{\z^{1} - 	\z^{1,\star} }^2 + C_{K,2} +  D_{K,2} ), \label{reg2}
	\end{align}
	where in (a) we have used the fact that for $\rho \in (0,1)$, it holds that $\frac{1}{(1-\rho^2)^2} \leq \frac{1}{(1-\rho)^2}$. Further, from Theorem~\ref{mytheo2} we have
	\begin{align}
		\Reg^K &\leq \frac{L \sigma \rho^2}{\lambda_{\min}(\P)(1-\rho)^2}(\Vert \z^1 - 	\z^{1, 	\star} \Vert_\P + W_K),\nonumber\\
		&\overset{(a)}{\leq} \frac{L \sqrt{\lambda_{\max}(\P)} \sigma_b 	\rho^2}{\lambda_{\min}(\P)(1-\rho)^2}( \sqrt{\lambda_{\max}(\P)} \Vert \z^1 -\z^{1, \star} \Vert +\sqrt{\lambda_{\max}(\P)} (C_K+ D_K)),\nonumber\\ 
		&= \frac{L \text{cond}(\P) \sigma_b \rho^2}{(1-\rho)^2}(  \Vert \z^1 -\z^{1, \star} \Vert +C_K+ D_K).\label{reg_comb}
	\end{align}
	where in (a) we have used the fact that $\norm{\z^{k,\star}-\z^{k-1,\star}}_{\P} \leq \sqrt{\lambda_{\max}(\P)} \norm{\z^{k,\star}-\z^{k-1,\star}} \leq \sqrt{\lambda_{\max}(\P)} \sigma_b$, and Lemma \ref{lem:wk_ck_dk}. Note that the bounds in Assumptions \eqref{a2}-\eqref{a3} translate to a bound of the form $\norm{\z^{k,\star}-\z^{k-1,\star}} \leq \sigma_b$. On combining \eqref{reg2} and \eqref{reg_comb}, we get the desired result.
	
	\footnotesize
	\bibliographystyle{IEEEtran}
	\bibliography{IEEEabrv,ref_up_final_v1}

\begin{thebibliography}{10}
\providecommand{\url}[1]{#1}
\csname url@samestyle\endcsname
\providecommand{\newblock}{\relax}
\providecommand{\bibinfo}[2]{#2}
\providecommand{\BIBentrySTDinterwordspacing}{\spaceskip=0pt\relax}
\providecommand{\BIBentryALTinterwordstretchfactor}{4}
\providecommand{\BIBentryALTinterwordspacing}{\spaceskip=\fontdimen2\font plus
\BIBentryALTinterwordstretchfactor\fontdimen3\font minus
  \fontdimen4\font\relax}
\providecommand{\BIBforeignlanguage}[2]{{%
\expandafter\ifx\csname l@#1\endcsname\relax
\typeout{** WARNING: IEEEtran.bst: No hyphenation pattern has been}%
\typeout{** loaded for the language `#1'. Using the pattern for}%
\typeout{** the default language instead.}%
\else
\language=\csname l@#1\endcsname
\fi
#2}}
\providecommand{\BIBdecl}{\relax}
\BIBdecl

\bibitem{nedic2009distributed}
A.~Nedic and A.~Ozdaglar, ``Distributed subgradient methods for multi-agent
  optimization,'' \emph{IEEE Transactions on Automatic Control}, vol.~54,
  no.~1, pp. 48--61, 2009.

\bibitem{ram2010distributed}
S.~S. Ram, A.~Nedi{\'c}, and V.~V. Veeravalli, ``Distributed stochastic
  subgradient projection algorithms for convex optimization,'' \emph{J. of
  Optim. Theory and Applications}, vol. 147, no.~3, pp. 516--545, 2010.

\bibitem{nedic2011asynchronous}
A.~Nedic, ``Asynchronous broadcast-based convex optimization over a network,''
  \emph{{IEEE} Trans. Autom. Control}, vol.~56, no.~6, pp. 1337--1351, 2011.

\bibitem{jakovetic2014fast}
D.~Jakoveti{\'c}, J.~Xavier, and J.~M. Moura, ``Fast distributed gradient
  methods,'' \emph{{IEEE} Trans. Autom. Control}, vol.~59, no.~5, pp.
  1131--1146, 2014.

\bibitem{yan2012distributed}
F.~Yan, S.~Sundaram, S.~Vishwanathan, and Y.~Qi, ``Distributed autonomous
  online learning: Regrets and intrinsic privacy-preserving properties,''
  \emph{{IEEE} Trans. Knowl. Data Eng.}, vol.~25, no.~11, pp. 2483--2493, 2012.

\bibitem{hosseini2013online}
S.~Hosseini, A.~Chapman, and M.~Mesbahi, ``Online distributed optimization via
  dual averaging,'' in \emph{IEEE CDC}, 2013, pp. 1484--1489.

\bibitem{lee2017stochastic}
S.~Lee, A.~Nedi{\'c}, and M.~Raginsky, ``Stochastic dual averaging for
  decentralized online optimization on time-varying communication graphs,''
  \emph{IEEE Transactions on Automatic Control}, vol.~62, no.~12, pp.
  6407--6414, 2017.

\bibitem{yi2020distributed}
X.~Yi, X.~Li, L.~Xie, and K.~H. Johansson, ``Distributed online convex
  optimization with time-varying coupled inequality constraints,'' \emph{IEEE
  Transactions on Signal Processing}, vol.~68, pp. 731--746, 2020.

\bibitem{oakamoto2023distributed}
K.~Oakamoto, N.~Hayashi, and S.~Takai, ``Distributed online adaptive gradient
  descent with event-triggered communication,'' \emph{IEEE Transactions on
  Control of Network Systems}, 2023.

\bibitem{yuan2016convergence}
K.~Yuan, Q.~Ling, and W.~Yin, ``On the convergence of decentralized gradient
  descent,'' \emph{SIAM J. on Optim.}, vol.~26, no.~3, pp. 1835--1854, 2016.

\bibitem{shi2015extra}
W.~Shi, Q.~Ling, G.~Wu, and W.~Yin, ``Extra: An exact first-order algorithm for
  decentralized consensus optimization,'' \emph{SIAM J. on Optim.}, vol.~25,
  no.~2, pp. 944--966, 2015.

\bibitem{li2019decentralized}
Z.~Li, W.~Shi, and M.~Yan, ``A decentralized proximal-gradient method with
  network independent step-sizes and separated convergence rates,''
  \emph{{IEEE} Trans. Signal Process.}, vol.~67, no.~17, pp. 4494--4506, 2019.

\bibitem{qu2017harnessing}
G.~Qu and N.~Li, ``Harnessing smoothness to accelerate distributed
  optimization,'' \emph{{IEEE} Trans. Control Netw. Syst.}, vol.~5, no.~3, pp.
  1245--1260, 2017.

\bibitem{nedic2017achieving}
A.~Nedic, A.~Olshevsky, and W.~Shi, ``Achieving geometric convergence for
  distributed optimization over time-varying graphs,'' \emph{SIAM J. on
  Optimization}, vol.~27, no.~4, pp. 2597--2633, 2017.

\bibitem{shahrampour2018distributed}
S.~Shahrampour and A.~Jadbabaie, ``Distributed online optimization in dynamic
  environments using mirror descent,'' \emph{{IEEE} Trans. Autom. Control},
  vol.~63, no.~3, pp. 714--725, 2018.

\bibitem{nazari2021adaptive}
P.~Nazari, E.~Khorram, and D.~A. Tarzanagh, ``Adaptive online distributed
  optimization in dynamic environments,'' \emph{Optimization Methods and
  Software}, vol.~36, no.~5, pp. 973--997, 2021.

\bibitem{zhang2019distributed}
Y.~Zhang, R.~J. Ravier, M.~M. Zavlanos, and V.~Tarokh, ``A distributed online
  convex optimization algorithm with improved dynamic regret,'' in \emph{IEEE
  CDC}, 2019, pp. 2449--2454.

\bibitem{zhang2020distributed}
Y.~Zhang, R.~J. Ravier, V.~Tarokh, and M.~M. Zavlanos, ``Distributed online
  convex optimization with improved dynamic regret,'' \emph{arXiv preprint
  arXiv:1911.05127}, 2020.

\bibitem{carnevale2022gtadam}
G.~Carnevale, F.~Farina, I.~Notarnicola, and G.~Notarstefano, ``{GTA}dam:
  Gradient tracking with adaptive momentum for distributed online
  optimization,'' \emph{IEEE Transactions on Control of Network Systems}, 2022.

\bibitem{alghunaim2022unified}
S.~A. Alghunaim and K.~Yuan, ``A unified and refined convergence analysis for
  non-convex decentralized learning,'' \emph{IEEE Transactions on Signal
  Processing}, vol.~70, pp. 3264--3279, 2022.

\bibitem{alghunaim2020decentralized}
S.~A. Alghunaim, E.~K. Ryu, K.~Yuan, and A.~H. Sayed, ``Decentralized proximal
  gradient algorithms with linear convergence rates,'' \emph{{IEEE} Trans.
  Autom. Control}, vol.~66, no.~6, pp. 2787--2794, 2020.

\bibitem{hong2017prox}
M.~Hong, D.~Hajinezhad, and M.-M. Zhao, ``Prox-{PDA}: The proximal primal-dual
  algorithm for fast distributed nonconvex optimization and learning over
  networks,'' in \emph{ICML}, 2017, pp. 1529--1538.

\bibitem{xu2015augmented}
J.~Xu, S.~Zhu, Y.~C. Soh, and L.~Xie, ``Augmented distributed gradient methods
  for multi-agent optimization under uncoordinated constant stepsizes,'' in
  \emph{IEEE CDC}, 2015, pp. 2055--2060.

\bibitem{scutari2019distributed}
G.~Scutari and Y.~Sun, ``Distributed nonconvex constrained optimization over
  time-varying digraphs,'' \emph{Math. Programming}, vol. 176, no. 1-2, pp.
  497--544, 2019.

\bibitem{sundararajan2017robust}
A.~Sundararajan, B.~Hu, and L.~Lessard, ``Robust convergence analysis of
  distributed optimization algorithms,'' in \emph{IEEE Allerton}, 2017, pp.
  1206--1212.

\bibitem{nedic2015decentralized}
A.~Nedi{\'c}, S.~Lee, and M.~Raginsky, ``Decentralized online optimization with
  global objectives and local communication,'' in \emph{IEEE ACC}, 2015, pp.
  4497--4503.

\bibitem{johansson2007simple}
B.~Johansson, M.~Rabi, and M.~Johansson, ``A simple peer-to-peer algorithm for
  distributed optimization in sensor networks,'' in \emph{IEEE CDC}, 2007, pp.
  4705--4710.

\bibitem{georgiadis2006resource}
L.~Georgiadis, M.~J. Neely, L.~Tassiulas \emph{et~al.}, ``Resource allocation
  and cross-layer control in wireless networks,'' \emph{Found. and Trends in
  Netw.}, vol.~1, no.~1, pp. 1--144, 2006.

\bibitem{mokhtari2016online}
A.~Mokhtari, S.~Shahrampour, A.~Jadbabaie, and A.~Ribeiro, ``Online
  optimization in dynamic environments: Improved regret rates for strongly
  convex problems,'' in \emph{IEEE CDC}, 2016, pp. 7195--7201.

\bibitem{zhang2017improved}
L.~Zhang, T.~Yang, J.~Yi, R.~Jin, and Z.-H. Zhou, ``Improved dynamic regret for
  non-degenerate functions,'' \emph{NeurIPS}, vol.~30, 2017.

\bibitem{bedi2018tracking}
A.~S. Bedi, P.~Sarma, and K.~Rajawat, ``Tracking moving agents via inexact
  online gradient descent algorithm,'' \emph{{IEEE} J. Sel. Topics Signal
  Process.}, vol.~12, no.~1, pp. 202--217, 2018.

\bibitem{zhang2018adaptive}
L.~Zhang, S.~Lu, and Z.-H. Zhou, ``Adaptive online learning in dynamic
  environments,'' \emph{NeurIPS}, vol.~31, 2018.

\bibitem{lesage2019online}
A.~Lesage-Landry, H.~Wang, I.~Shames, P.~Mancarella, and J.~A. Taylor, ``Online
  convex optimization of multi-energy building-to-grid ancillary services,''
  \emph{{IEEE} Trans. Control Syst. Technol.}, vol.~28, no.~6, pp. 2416--2431,
  2019.

\bibitem{dixit2019online}
R.~Dixit, A.~S. Bedi, R.~Tripathi, and K.~Rajawat, ``Online learning with
  inexact proximal online gradient descent algorithms,'' \emph{{IEEE} Trans.
  Signal Process.}, vol.~67, no.~5, pp. 1338--1352, 2019.

\bibitem{chiang2012online}
C.-K. Chiang, T.~Yang, C.-J. Lee, M.~Mahdavi, C.-J. Lu, R.~Jin, and S.~Zhu,
  ``Online optimization with gradual variations,'' in \emph{Conference on
  Learning Theory}.\hskip 1em plus 0.5em minus 0.4em\relax JMLR Workshop and
  Conference Proceedings, 2012, pp. 6--1.

\bibitem{chiang2013beating}
C.-K. Chiang, C.-J. Lee, and C.-J. Lu, ``Beating bandits in gradually evolving
  worlds,'' in \emph{Conference on Learning Theory}.\hskip 1em plus 0.5em minus
  0.4em\relax PMLR, 2013, pp. 210--227.

\bibitem{matei2011performance}
I.~Matei and J.~S. Baras, ``Performance evaluation of the consensus-based
  distributed subgradient method under random communication topologies,''
  \emph{{IEEE} J. Sel. Topics Signal Process.}, vol.~5, no.~4, pp. 754--771,
  2011.

\bibitem{bianchi2015coordinate}
P.~Bianchi, W.~Hachem, and F.~Iutzeler, ``A coordinate descent primal-dual
  algorithm and application to distributed asynchronous optimization,''
  \emph{{IEEE} Trans. Autom. Control}, vol.~61, no.~10, pp. 2947--2957, 2015.

\bibitem{duchi2011dual}
J.~C. Duchi, A.~Agarwal, and M.~J. Wainwright, ``Dual averaging for distributed
  optimization: Convergence analysis and network scaling,'' \emph{{IEEE} Trans.
  Autom. Control}, vol.~57, no.~3, pp. 592--606, 2011.

\bibitem{boyd2011distributed}
S.~Boyd, N.~Parikh, E.~Chu, B.~Peleato, J.~Eckstein \emph{et~al.},
  ``Distributed optimization and statistical learning via the alternating
  direction method of multipliers,'' \emph{Found. and Trends in Machine
  Learning}, vol.~3, no.~1, pp. 1--122, 2011.

\bibitem{iutzeler2015explicit}
F.~Iutzeler, P.~Bianchi, P.~Ciblat, and W.~Hachem, ``Explicit convergence rate
  of a distributed alternating direction method of multipliers,'' \emph{{IEEE}
  Trans. Autom. Control}, vol.~61, no.~4, pp. 892--904, 2015.

\bibitem{ling2015dlm}
Q.~Ling, W.~Shi, G.~Wu, and A.~Ribeiro, ``{DLM}: Decentralized linearized
  alternating direction method of multipliers,'' \emph{{IEEE} Trans. Signal
  Process.}, vol.~63, no.~15, pp. 4051--4064, 2015.

\bibitem{chen2012fast}
A.~I. Chen and A.~Ozdaglar, ``A fast distributed proximal-gradient method,'' in
  \emph{IEEE Allerton}, 2012, pp. 601--608.

\bibitem{tsitsiklis1986distributed}
J.~Tsitsiklis, D.~Bertsekas, and M.~Athans, ``Distributed asynchronous
  deterministic and stochastic gradient optimization algorithms,'' \emph{{IEEE}
  Trans. Autom. Control}, vol.~31, no.~9, pp. 803--812, 1986.

\bibitem{nedic2018distributed}
A.~Nedi{\'c} and J.~Liu, ``Distributed optimization for control,'' \emph{Annual
  Review of Control, Robotics, and Autonomous Systems}, vol.~1, pp. 77--103,
  2018.

\bibitem{sun2017distributed}
Y.~Sun and G.~Scutari, ``Distributed nonconvex optimization for sparse
  representation,'' in \emph{IEEE ICASSP}, 2017, pp. 4044--4048.

\bibitem{lu2019gnsd}
S.~Lu, X.~Zhang, H.~Sun, and M.~Hong, ``{GNSD}: A gradient-tracking based
  nonconvex stochastic algorithm for decentralized optimization,'' in
  \emph{IEEE DSW}, 2019, pp. 315--321.

\bibitem{chang2020distributed}
T.-H. Chang, M.~Hong, H.-T. Wai, X.~Zhang, and S.~Lu, ``Distributed learning in
  the nonconvex world: From batch data to streaming and beyond,'' \emph{{IEEE}
  Signal Process. Mag.}, vol.~37, no.~3, pp. 26--38, 2020.

\bibitem{tang2018d}
H.~Tang, X.~Lian, M.~Yan, C.~Zhang, and J.~Liu, ``{D2}: Decentralized training
  over decentralized data,'' in \emph{ICML}, 2018, pp. 4848--4856.

\bibitem{wu2022decentralized}
W.~Wu, Z.~Li, Y.~Zhao, C.~Yu, P.~Zhao, J.~Liu, and K.~He, ``Decentralized
  online learning: Take benefits from others’ data without sharing your own
  to track global trend,'' \emph{ACM Trans. on Intell. Systems and Technol.},
  vol.~14, no.~1, pp. 1--22, 2022.

\bibitem{dixit2020online}
R.~Dixit, A.~S. Bedi, and K.~Rajawat, ``Online learning over dynamic graphs via
  distributed proximal gradient algorithm,'' \emph{{IEEE} Trans. Autom.
  Control}, vol.~66, no.~11, pp. 5065--5079, 2020.

\bibitem{nazari2022dadam}
P.~Nazari, D.~A. Tarzanagh, and G.~Michailidis, ``{DA}dam: A consensus-based
  distributed adaptive gradient method for online optimization,'' \emph{{IEEE}
  Trans. Signal Process.}, 2022.

\bibitem{lesage2020dynamic}
A.~Lesage-Landry and D.~S. Callaway, ``Dynamic and distributed online convex
  optimization for demand response of commercial buildings,'' \emph{IEEE
  Control Sys. Lett.}, vol.~4, no.~3, pp. 632--637, 2020.

\bibitem{eshraghi2022improving}
N.~Eshraghi and B.~Liang, ``Improving dynamic regret in distributed online
  mirror descent using primal and dual information,'' in \emph{Learning for
  Dynamics and Control Conference}.\hskip 1em plus 0.5em minus 0.4em\relax
  PMLR, 2022, pp. 637--649.

\bibitem{derenick2009optimal}
J.~Derenick, J.~Spletzer, and A.~Hsieh, ``An optimal approach to collaborative
  target tracking with performance guarantees,'' \emph{Journal of Intelligent
  and Robotic Systems}, vol.~56, pp. 47--67, 2009.

\bibitem{derenick2009convex}
J.~C. Derenick, \emph{A convex optimization framework for multi-agent motion
  planning}.\hskip 1em plus 0.5em minus 0.4em\relax Lehigh University, 2009.

\bibitem{candanedo2016accurate}
L.~M. Candanedo and V.~Feldheim, ``Accurate occupancy detection of an office
  room from light, temperature, humidity and co2 measurements using statistical
  learning models,'' \emph{Energy and Buildings}, vol. 112, pp. 28--39, 2016.

\bibitem{freund2018condition}
R.~M. Freund, P.~Grigas, and R.~Mazumder, ``Condition number analysis of
  logistic regression, and its implications for standard first-order solution
  methods,'' \emph{arXiv preprint arXiv:1810.08727}, 2018.

\end{thebibliography}
	
	\newcounter{mainlemma}
	\setcounter{mainlemma}{\value{lemma}} 
	
\end{document}


\title{Supplementary Material \author{Shivangi Dubey Sharma and Ketan Rajawat, \textit{Member, IEEE}}
	}
	
	\maketitle
	\setcounter{lemma}{3}
	\setcounter{figure}{6}
	\setcounter{equation}{58}
	\section{Proof of Lemma 1}
	In this section, we establish the modified co-coercivity property of the $L_i$-smooth and $\M_i$-convex function $f_i$. We will begin with stating few preliminary results that are required for proof.	
	
	\begin{lemma} \label{lem:ConvexityOfh}
		If $f(\x)$ is a $L$-smooth convex function with $g(\x) = f(\x) - \frac{1}{2} \x^\mathsf{T} \M \x$ being convex, then $$h(\x)= \frac{1}{2} \x^\mathsf{T} (L \I - \M) \x - g(\x),$$ is a convex function.
	\end{lemma} 
	\begin{IEEEproof}
		From the definition of $g(\x)$ we have 
		\begin{align}
			h(\x) &= \frac{1}{2} \x^\mathsf{T} (L \I - \M) \x - f(\x) + \frac{1}{2} \x^\mathsf{T} \M \x \\	
			&= \frac{L}{2} \x^\mathsf{T}\x - f(\x)
		\end{align}
		which is convex since $f$ is an $L$-smooth convex function.	
	\end{IEEEproof}
	
	\begin{lemma} \label{lem:quadrUpper}
		If $h(\x)= \frac{1}{2} \x^\mathsf{T} (L \I - \M) \x - g(\x)$ is a convex function, where $L \I - \M$ is a symmetric matrix, then $$g(\y) \leq g(\x) + \nabla g(\x)^\mathsf{T} (\y -\x) + \frac{1}{2} (\y -\x)^\mathsf{T} (L\I - \M)(\y -\x),$$
		for all $\x$ and $\y$ in domain of $g$ 
	\end{lemma} 
	\begin{IEEEproof}
		From the first order convexity condition of $h(\x)$ we have,
		\begin{align*}
			h(\y) &\geq h(\x) + \nabla h(\x)^\T(\y - \x), \\
			\Rightarrow \y^\T\left(\frac{L\I - \M}{2} \right)\y - g(\y) &\geq  	\x^\T \left(\frac{L\I - \M}{2} \right)\x-g(\x) + ((L\I - \M)\x - \nabla g(\x))^\T (\y - \x). 
		\end{align*} 
		which upon simplifying yields the desired result.
	\end{IEEEproof}
	
	\begin{lemma} \label{lem:lowerbound}
		If $h(\x)= \frac{1}{2} \x^\mathsf{T} (L \I - \M) \x - g(\x)$ is a convex function, where $L \I - \M$ is an invertible symmetric matrix and $\x^\star$ is the minimizer of $g$ then
		$$g(\x) -g(\x^{\star}) \geq \frac{1}{2}g(\x) + \nabla g(\x)^\mathsf{T} (L \I - \M)^{-1}\nabla g(\x).$$ 
	\end{lemma} 
	\begin{IEEEproof}
		From the Lemma \ref{lem:quadrUpper} we have,
		\begin{align*}
			g(\y) \leq g(\x) + \nabla g(\x)^\mathsf{T} (\y -\x) + \frac{1}{2} (\y -\x)^\mathsf{T} (L\I - \M)(\y -\x),
		\end{align*} 
		As $\x^\star$ is the minimizer of $g(\x)$, we have
		\begin{align}
			g(\x^\star) = \min_\y g(\y) &\leq \underset{\y }{\min}\{ g(\x) + \nabla g(\x)^\mathsf{T} (\y -\x) + \frac{1}{2} (\y -\x)^\mathsf{T} (L\I - \M)(\y -\x)\}. \label{ming}
		\end{align}
		Observe that $\y = \x - (L \I -\M)^{-1} \nabla g(\x)$ minimizes the right-hand side of \ref{ming}, so
		\begin{align*}
			g(\x^\star) &\leq g(\x) - \nabla g(\x)^\T (L \I -\M)^{-1} \nabla g(\x) + \nabla g(\x)^\T (L \I -\M)^{-1} \left(\frac{L\I - \M}{2} \right) (L \I -\M)^{-1} \nabla g(\x).
		\end{align*}
		On simplifying further, we obtain the desired result.
	\end{IEEEproof}
	
	\begin{lemma} \label{lem:co-coercivity}
		If $g(\x)$ is convex and $h(\x)= \frac{1}{2} \x^\mathsf{T} (L \I - \M) \x - g(\x)$ is a convex function, where $L \I - \M$ is an invertible symmetric matrix then
		\begin{align*}
			(\nabla g(\x) - \nabla g(\y))^\mathsf{T} (\x - \y) \geq (\nabla g(\x) - \nabla g(\y))^\mathsf{T} (L \I - \M)^{-1}(\nabla g(\x) - \nabla g(\y)) .
		\end{align*} 
	\end{lemma} 
	\begin{IEEEproof}
		Let us define  $g_\x(\z) = g(\z) - \nabla g(\x)^\T \z $ and $g_\y(\z) = g(\z) - \nabla g(\y)^\T \z $. As the sum of convex functions is also convex, it holds that $\z^\T \left(\frac{L \I - \M}{2}\right) \z - g_\x(\z)$ and  $\z^\T \left(\frac{L \I - \M}{2}\right) \z - g_\y(\z)$ are convex in $\z$. Also, $\z=\x$ minimizes $g_\x(\z)$, so from Lemma \ref{lem:lowerbound} we obtain
		\begin{align*}
			g_{\x}(\y) - g_{\x}(\x) \geq \frac{1}{2} \nabla g_{\x}(\y)^\T (L\I-\M)^{-1} \nabla g_{\x}(\y).
		\end{align*}
		From the definition of $g_{\x}(\y)$ we have $g_{\x}(\y) - g_{\x}(\x) = g(\y)-g(\x) + \nabla g(\x)^\T(\x - \y)$. Hence,
		\begin{align}
			g(\y) -g(\x) + \nabla g(\x)^\T(\x - \y) &\geq \frac{1}{2} \nabla g_{\x}(\y)^\T (L\I-\M)^{-1} \nabla g_{\x}(\y) \label{eq1}\\
			&= \frac{1}{2} (\nabla g(\y) - \nabla g(\x))^\T (L\I-\M)^{-1} (\nabla g(\y) - \nabla g(\x)). \nonumber
		\end{align}
		Similarly $\z = \y$ minimizes $g_{\y}(\z)$ and we have
		\begin{align}
			&g(\x) -g(\y) + \nabla g(\y)^\T(\x - \y) \geq  \frac{1}{2} (\nabla g(\x) - \nabla g(\y))^\T (L\I-\M)^{-1} (\nabla g(\x) - \nabla g(\y)). \label{eq2}
		\end{align}
		On adding \eqref{eq1} and \eqref{eq2}, we get the desired result.
	\end{IEEEproof}
	
	Having established the preliminary lemmas, we are now ready to prove Lemma 1. 
	\begin{IEEEproof}[Proof of Lemma 1]
		As $f_i(\x)$ is $L_i$-smooth and $g(\x) = f_i(\x) - \x^\mathsf{T} \M_i \x$ is convex, it follows from Lemma \ref{lem:ConvexityOfh} that $h(\x)= \frac{1}{2} \x^\mathsf{T} (L_i \I - \M_i) \x - g(\x)$ is convex. Therefore from Lemma \ref{lem:co-coercivity}, we have
		\begin{align*}
			(\nabla g(\x) - \nabla g(\y))^\mathsf{T} (\x - \y) \geq (\nabla g(\x) - \nabla g(\y))^\mathsf{T} (L_i \I - \M_i)^{-1}(\nabla g(\x) - \nabla g(\y)) .
		\end{align*}
		Substituting the definition of $g(\x)$ we get
		\begin{align*}
			&(\nabla f_i(\x) - \M_i \x - \nabla f_i(\y) + \M_i \y)^\mathsf{T} (\x - \y) \geq \\ 
			&(\nabla f_i(\x) - \M_i \x - \nabla f_i(\y) + \M_i \y)^\mathsf{T} (L_i \I - \M_i)^{-1}(\nabla f(\x) - \M_i \x - \nabla f(\y) + \M_i \y) .
		\end{align*}
		After further simplification, we obtain
		\begin{align*}
			&(\nabla f_i(\x) -  \nabla f_i(\y))^\mathsf{T}(\x - \y) + (\nabla f_i(\x) - \nabla f_i(\y))^\mathsf{T}	(L_i \I - \M_i)^{-1} \M_i(\x -\y)) \\ 
			&+(\nabla f(\x) - \nabla f(\y))  ^\mathsf{T} \M(L_i\I - \M_i)^{-1}(\x - \y) \\
			&\geq(\nabla f_i(\x) - \nabla f_i(\y))^\mathsf{T} (L_i \I - \M_i)^{-1}(\nabla f_i(\x)-  \nabla f_i(\y))(\x - \y)^\mathsf{T} \M_i (\x - \y) \\
			&+ (\x - \y)^\mathsf{T}\M_i (L_i \I - \M_i)^{-1} \M_i(\x -\y).
		\end{align*}
		Multiplying with 2 and writing in vector form, we obtain
		\begin{align}
			\begin{bmatrix} \x- \y \\ \nabla f_i(\x) - \nabla f_i(\y) \end{bmatrix}^{\mathsf{T}} \S \begin{bmatrix} \x- \y\\ \nabla f_i(\x) - \nabla f_i(\y) \end{bmatrix}\geq 0 \label{genco}
		\end{align} 
		where 
		\begin{align}
			\S = \begin{bmatrix} -2(\M_i(L_i\I - \M_i)^{-1}\M_i+\M_i)  &\I+2(L_i\I - \M_i)^{-1}\M_i \\  \I+2(L_i\I - \M_i)^{-1}\M_i  & -2(L_i\I - \M_i)^{-1} \end{bmatrix} \nonumber
		\end{align} 
		which can be further simplified to 
		\begin{align}
			\S =\begin{bmatrix} -2L_i(L_i\I - \M_i)^{-1}\M_i  &(L_i\I - \M_i)^{-1}(L_i\I + \M_i) \\  
				(L_i\I - \M_i)^{-1}(L_i\I+\M_i)  & -2(L_i\I - \M_i)^{-1} \end{bmatrix}. \nonumber
		\end{align} 
		The desired result then follows from substituting $\S$ in \eqref{genco} and simplifying.  
	\end{IEEEproof}

	\section{Additional Simulation Results}
	In this section, we include addition simulation results that demonstrate the robustness of the proposed algorithm across different target variability, captured in the target tracking example (Sec.V-A) by the frequency of the sinusoid. 
	\begin{figure*}[htb]
		\centering
		\includegraphics[scale=0.4]{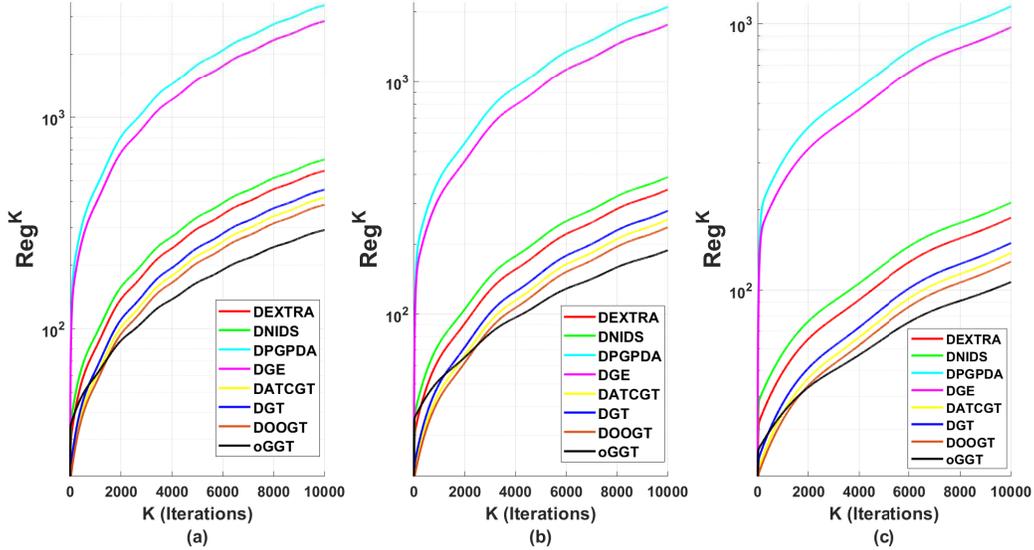}
		\caption{Dynamic Regret Rate of D-EXTRA, D-NIDS, D-PGPDA, D-GE, D-ATC-GT, D-GT, DOOGT and oGGT in various cases where target signal varies with angular frequency (a) 6.66$\pi$ , (b) 5$\pi$, and (c) 2$\pi$}
		\label{fig:regret}
	\end{figure*}
	It can be seen from Fig. \ref{fig:regret} that oGGT outperforms all the other algorithms across all frequencies. Indeed, the relative performance of various algorithms remains same across target frequencies.